\documentclass[10pt,twocolumn,letterpaper]{article}

\usepackage{cvpr}
\usepackage{times,bm,pifont}
\usepackage{graphicx}
\usepackage{amsmath}
\usepackage{amssymb}

\usepackage[american]{babel}
\usepackage{multirow}
\usepackage{subcaption}
\usepackage{xcolor}
\usepackage{algorithm}
\usepackage{algorithmicx}
\usepackage{algpseudocode}

\usepackage{caption}
\newcommand{\cmark}{\ding{51}}%
\newcommand{\xmark}{\ding{55}}%

\newtheorem{proposition}{Proposition}
\newtheorem{observation}{Observation}

\usepackage[breaklinks=true,bookmarks=false,colorlinks]{hyperref}

\cvprfinalcopy 

\def\cvprPaperID{2818} 

\ifcvprfinal\fi
\begin{document}

\title{Towards Accurate One-Stage Object Detection with AP-Loss}

\author{\small Kean Chen$^1$, Jianguo Li$^2$, Weiyao Lin$^{1}$\!\thanks{Corresponding Author, Email: wylin@sjtu.edu.cn}\,\,, John See$^3$, Ji Wang$^4$, Lingyu Duan$^5$, Zhibo Chen$^4$, Changwei He$^4$, Junni Zou$^1$\\
        \small $^{1}$Shanghai Jiao Tong University, China, $^{2}$ Intel Labs, China, \\
        \small $^{3}$ Multimedia University, Malaysia, $^{4}$ Tencent YouTu Lab, China, $^{5}$ Peking University, China
}

\maketitle
\thispagestyle{empty}

\begin{abstract}
    One-stage object detectors are trained by optimizing classification-loss and localization-loss simultaneously, with the former suffering much from extreme foreground-background class imbalance issue due to the large number of anchors. This paper alleviates this issue by proposing a novel framework to replace the classification task in one-stage detectors with a ranking task, and adopting the Average-Precision loss (AP-loss) for the ranking problem. Due to its non-differentiability and non-convexity, the AP-loss cannot be optimized directly. For this purpose, we develop a novel optimization algorithm, which seamlessly combines the error-driven update scheme in perceptron learning and backpropagation algorithm in deep networks. We verify good convergence property of the proposed algorithm theoretically and empirically. Experimental results demonstrate notable performance improvement in state-of-the-art one-stage detectors based on AP-loss over different kinds of classification-losses on various benchmarks, without changing the network architectures. Code is available at \url{https://github.com/cccorn/AP-loss}.
\end{abstract}

\section{Introduction}

\begin{figure}[t]
\small
\begin{subfigure}[b]{1.0\linewidth}
  \centering
  \includegraphics[width=0.95\linewidth]{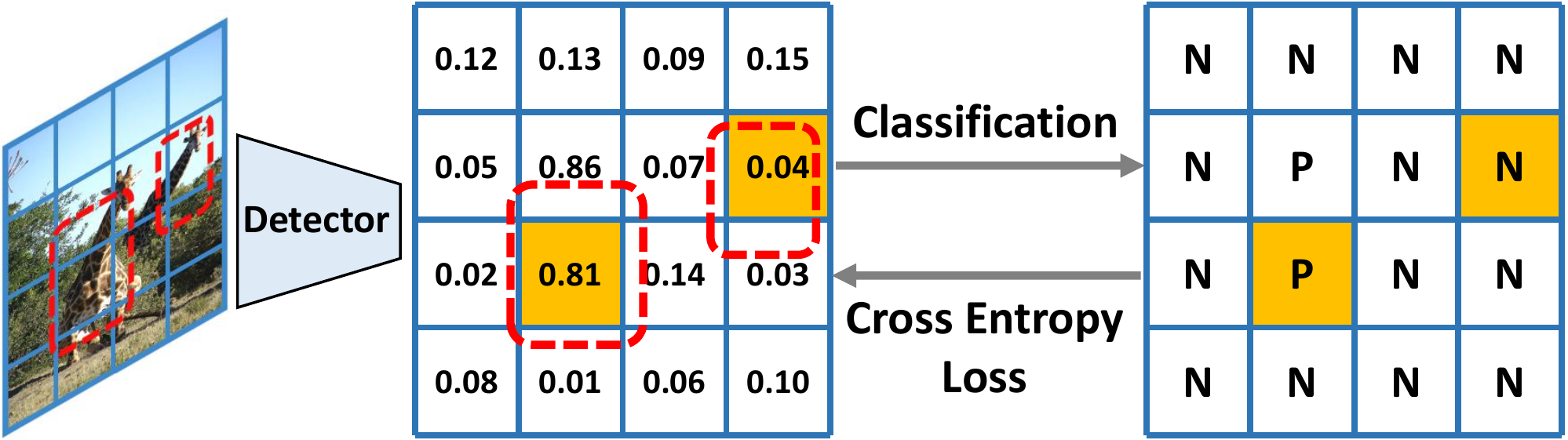}
  \vspace{-1ex}
  \caption{Acc \(=0.88\)}
  \label{fig-classification}
\end{subfigure}
\par\smallskip
\begin{subfigure}[b]{1.0\linewidth}
  \centering
  \includegraphics[width=0.95\linewidth]{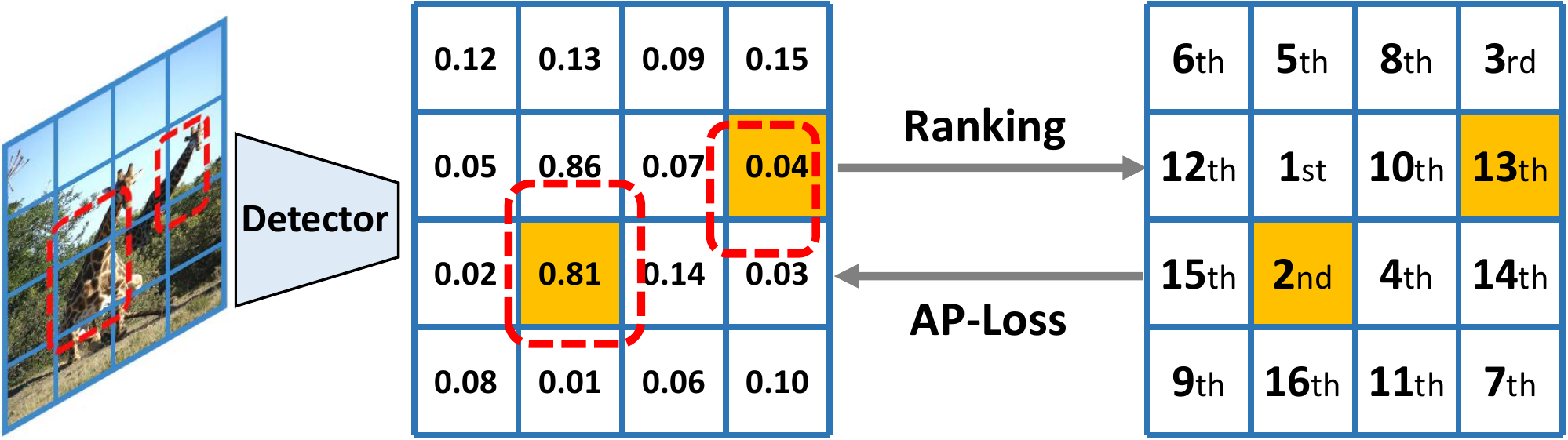}
  \vspace{-1ex}
  \caption{AP \(=0.33\)}
  \label{fig-ranking}
\end{subfigure}
\vspace{-5ex}
  \caption{Dashed red boxes are the ground truth object boxes. The orange filled boxes and other blank boxes are anchors with positive and negative ground truth labels, repectively. (a) shows that the detection performance is poor but the classification accuracy is still high due to large number of true negatives. (b) shows the ranking metric AP can better reflect the actual condition as it does not suffer from the large number of true negatives.}
\vspace{-4ex}
\end{figure}

\begin{figure*}[ht]
\centering
\small
  \includegraphics[width=0.88\linewidth]{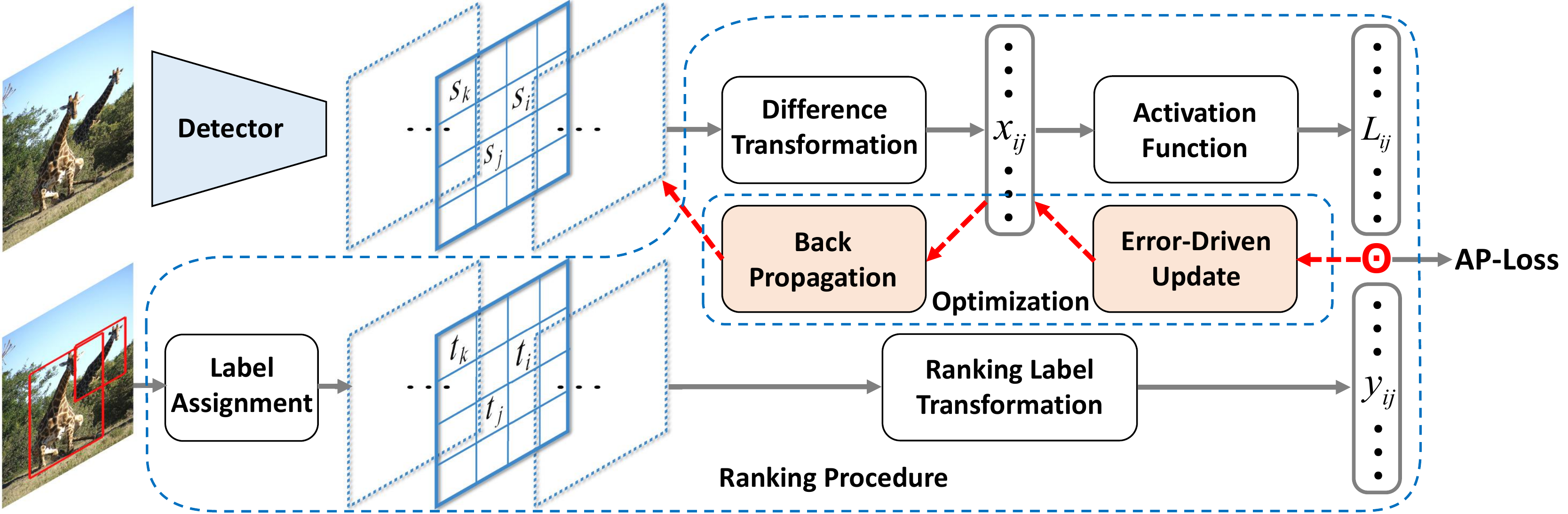}
\vspace{-2.5mm}
  \caption{Overall framework of the proposed approach. We replace the classification-task in one-stage detectors with a ranking task, where the ranking procedure produces the primary terms of AP-loss and the corresponding label vector. The optimization algorithm is based on an error-driven learning scheme combined with backpropagation. The localization-task branch is not shown here due to no modification.}
\vspace{-4.5mm}
\label{fig-overall}
\end{figure*}

Object detection needs to localize and recognize the objects simultaneously from the large backgrounds, which remains challenging due to the imbalance between foreground and background.
Deep learning based detection solutions usually adopt a multi-task architecture, which handles classification task and localization task with different loss functions.
The classification task aims to recognize the object in a given box, while the localization task aims to predict the precise bounding box of the object. Two-stage detectors~\cite{ren2015faster,girshick2015fast,dai2016r,lin2017feature} first generates a limited number of object box proposals, so that the detection problem can be solved by adopting classification task on those proposals. However, the circumstance is different for one-stage detectors, which need to predict the object class directly from the densely pre-designed candidate boxes. The large number of boxes yield the imbalance between foreground and background which makes the optimization of classification task easily biased and thus impacts the detection performance. It is observed that the classification metric could be very high for a trivial solution which predicts negative label for almost all candidate boxes, while the detection performance is poor. \autoref{fig-classification} illustrates one such example.

To tackle this issue in one-stage object detectors, some works introduce new classification losses such as balanced loss~\cite{redmon2016you}, Focal Loss~\cite{lin2018focal}, as well as tailored training method such as Online Hard Example Mining (OHEM)~\cite{liu2016ssd,shrivastava2016ohem}. These losses model each sample (anchor box) independently, and attempt to re-weight the foreground and background samples in classification losses to cater for the imbalance condition; this is done without considering the relationship among different samples. The designed balance weights are hand-crafted hyper-parameters, which do not generalize well across datasets. We argue that the gap between classification task and detection task hinder the performance of one-stage detectors. In this paper, instead of modifying the classification loss, we propose to replace classification task with ranking task in one-stage detectors, where the associated ranking loss explicitly models sample relationship, and is invariant to the ratio of positive and negative samples. As shown in \autoref{fig-ranking}, we adopt Average Precision (AP) as our target loss which is inherently more consistent with the evaluation metric for object detection.

However, it is non-trivial to directly optimize the AP-loss due to the non-differentiability and non-decomposability, so that
standard gradient descent methods are not amenable for this case.
There are three aspects of studies for this issue.
\textit{First}, AP based loss is studied within structured SVM models~\cite{yue2007support,mohapatra2014efficient}, which restricts in linear SVM model so that the performance is limited.
\textit{Second}, a structured hinge loss~\cite{Mohapatra_2018_CVPR} is proposed to optimize the upper bound of AP-loss instead of the loss itself.
\textit{Third}, approximate gradient methods~\cite{song2016training,henderson2016end} are proposed for optimizing the AP-loss, which are less efficient and easy to fall into local optimum even for the case of linear models due to the non-convexity and non-quasiconvexity of the AP-loss.
Therefore, it is still an open problem for the optimization of the AP-loss.

In this paper, we address this challenge by replacing the classification task in one-stage detectors with a ranking task,
so that we handle the class imbalance problem with a ranking based loss named AP-loss. Furthermore, we propose a novel error-driven learning algorithm to effectively optimize the non-differentiable AP based objective function.
More specifically, some extra transformations are added to the score output of one-stage detector to obtain the AP-loss, which includes a linear transformation that transforms the scores to pairwise differences, and a non-linear and non-differentiable ``activation function'' that transform the pairwise differences to the primary terms of the AP-loss. Then the AP-loss can be obtained by the dot product between the primary terms and the label vector.
It is worth noting that the difficulty for using gradient method on the AP-loss lies in passing gradients through the non-differentiable activation function.
Inspired by the perceptron learning algorithm~\cite{rosenblatt1957perceptron}, we adopt an error-driven learning scheme to directly pass the update signal through the non-differentiable activation function. Different from gradient method, our learning scheme gives each variable an update signal proportional to the error it makes.
Then, we adopt the backpropagation algorithm to transfer the update signal to the weights of neural network.
We theoretically and experimentally prove that the proposed optimization algorithm does not suffer from the non-differentiability and non-convexity of the objective function. The main contributions of this paper are summarized as below:
\begin{itemize}
\vspace{-0mm}
\setlength{\topsep}{0pt}
\setlength{\itemsep}{1pt}
\setlength{\parskip}{1pt}
\item We propose a novel framework in one-stage object detectors which adopts the ranking loss to handle the class imbalance issue.
\item We propose an error-driven learning algorithm that can efficiently optimize the non-differentiable and non-convex AP-based objective function with both theoretical and experimental verifications.
\item We show notable performance improvement with the proposed method on state-of-the-art one-stage detectors over different kinds of classification-losses without changing the model architecture.
\end{itemize}

\section{Related Work}

{\bf \noindent One-stage detectors:}
In object detection, the one-stage approaches have relatively simpler architecture and higher efficiency than two-stage approaches. OverFeat~\cite{sermanet2013overfeat} is one of the first CNN-based one-stage detectors. Thereafter, different designs of one-stage detectors are proposed, including SSD~\cite{liu2016ssd}, YOLO~\cite{redmon2016you}, DSSD~\cite{fu2017dssd} and DSOD~\cite{shen2017dsod,li2018tiny}. These methods demonstrate good processing efficiency as one-stage detectors, but generally yield lower accuracy than two-stage detectors.
Recently, RetinaNet~\cite{lin2018focal} and RefineDet~\cite{zhang2018single} narrow down the performance gap (especially on the challenging COCO benchmark~\cite{lin2014microsoft}) between one-stage approaches and two-stage approaches with some innovative designs.
As commonly known, the performance of one-stage detectors benefits much from densely designed anchors, which introduce extreme imbalance between foreground and background samples. To address this challenge, methods like OHEM~\cite{liu2016ssd,shrivastava2016ohem} and Focal Loss~\cite{lin2018focal} have been proposed to reduce the loss weight for easy samples.
However, there are two hurdles that are still open to discussion. Firstly, hand-crafted hyper-parameters for weight balance do not generalize well across datasets. Secondly, the relationship among sample anchors is far from well modeled.

{\bf \noindent AP as a loss for Object Detection:}
Average Precision (AP) is widely used as the evaluation metric in many tasks such as object detection~\cite{everingham2015pascal} and information retrieval~\cite{salton1986introduction}.
However, AP is far from a good and common choice as an optimization goal in object detection due to its non-differentiability and non-convexity.
Some methods have been proposed to optimize the AP-loss in object detection, such as AP-loss in the linear structured SVM model~\cite{yue2007support,mohapatra2014efficient}, structured hinge loss as upper bound of the AP-loss~\cite{Mohapatra_2018_CVPR}, approximate gradient methods~\cite{song2016training,henderson2016end},
reinforcement learning to fine-tune a pre-trained object detector with AP based metric~\cite{rao2018learning}.
Although these methods give valuable results in optimizing the AP-loss, their performances are still limited due to the intrinsic limitations.
In details, the proposed approach differs from them in 4 aspects.
(1) Our approach can be used for any differentiable linear or non-linear models such as neural networks, while \cite{yue2007support,mohapatra2014efficient} only work for linear SVM model.
(2) Our approach directly optimizes the AP-loss, while \cite{Mohapatra_2018_CVPR} introduces notable loss gap after relaxation.
(3) Our approach dose not approximate the gradient and dose not suffer from the non-convexity of objective function as in \cite{song2016training,henderson2016end}.
(4) Our approach can train the detectors in an end-to-end way, while \cite{rao2018learning} cannot.

{\bf \noindent Perceptron Learning Algorithm:}
The core of our optimization algorithm is the ``error-driven update'' which is generalized from the perceptron learning algorithm~\cite{rosenblatt1957perceptron}, and helps overcome the difficulty of the non-differentiable objective functions. The perceptron is a simple artificial neuron using the Heaviside step function as the activation function.
The learning algorithm was first invented by Frank Rosenblatt~\cite{rosenblatt1957perceptron}. As the Heaviside step function in perceptron is non-differentiable, it is not amenable for gradient method.
Instead of using a surrogate loss like cross-entropy, the perceptron learning algorithm employs an error-driven update scheme directly on the weights of neurons. This algorithm is guaranteed to converge in finite steps if the training data is linearly separable.
Further works like~\cite{krauth1987learning,anlauf1989adatron,wendemuth1995learning} have studied and improved the stability and robustness of the perceptron learning algorithm.

\section{Method}
We aim to replace the classification task with AP-loss based ranking task in one-stage detectors such as RetinaNet~\cite{lin2018focal}.
\autoref{fig-overall} shows the two key components of our approach, \textit{i.e.}, the ranking procedure and the error-driven optimization algorithm.
Below, we will first present how AP-loss is derived from traditional score output.
Then, we will introduce the error-driven optimization algorithm.
Finally, we also present the theoretical analyses of the proposed optimization algorithm and outline the training details.
Note that all changes are made on the loss part of the classification branch without changing the backbone model and localization branch.

\subsection{Ranking Task and AP-Loss}
\subsubsection{Ranking Task}

\begin{figure}[t]
\centering
\small
\begin{subfigure}[b]{0.3\linewidth}
  \centering
  \includegraphics[width=1\linewidth]{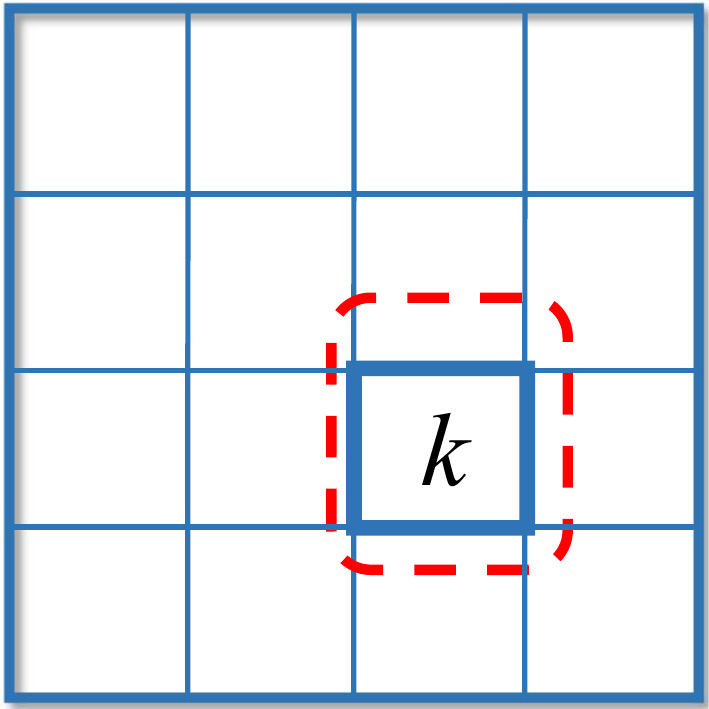}
  \vspace{-3.5ex}
  \caption{}
\end{subfigure}
\quad
\begin{subfigure}[b]{0.5\linewidth}
  \centering
  \includegraphics[width=1\linewidth]{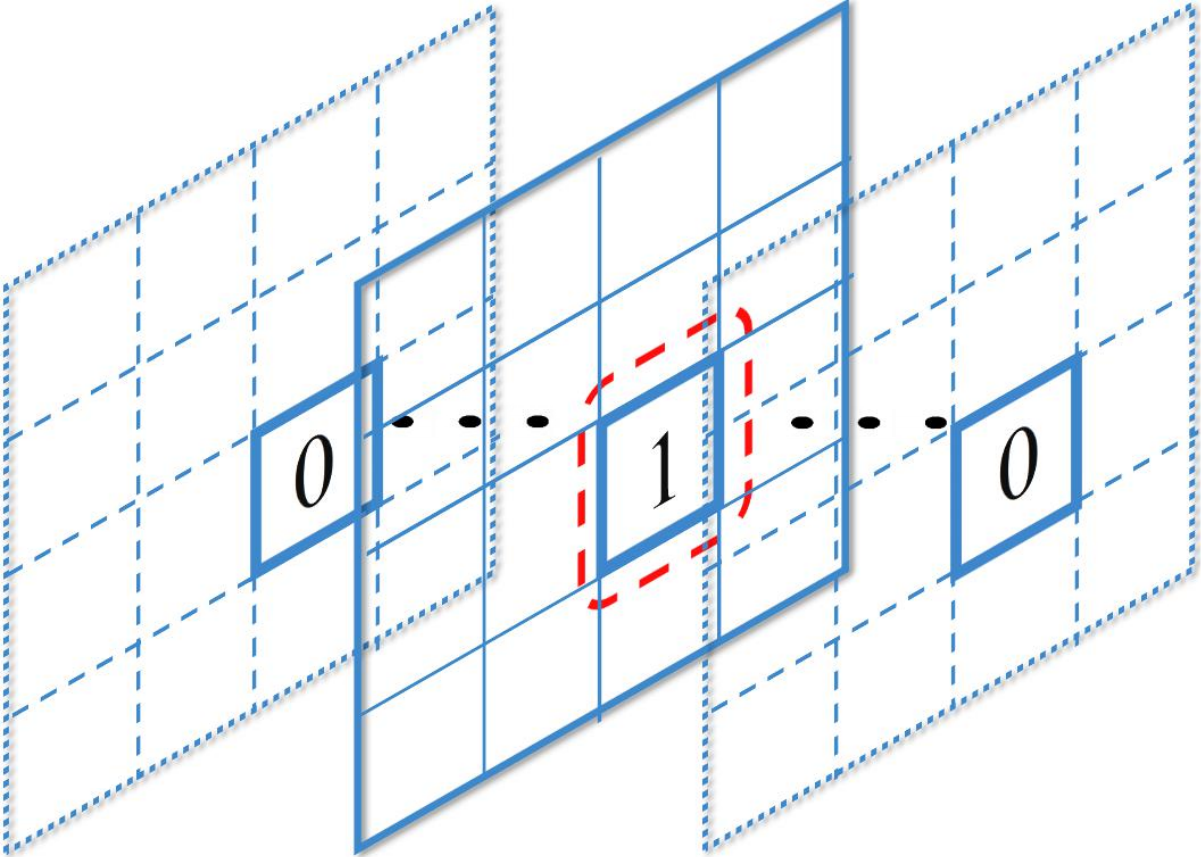}
  \vspace{-3.5ex}
  \caption{}
\end{subfigure}
\vspace{-2ex}
  \caption{Comparison of label assignments. The dashed red box is the ground truth box with class $k$.
  (a) In traditional classification task of one-stage detectors, the anchor is assigned a foreground label $k$.
  (b) In our ranking task framework, the anchor replicates $K$ times, and we assign the $k$-th anchor to label $1$, others 0.}
\label{fig-labelassignment}
\vspace{-3ex}
\end{figure}

In traditional one-stage detectors, given input image $I$, suppose the pre-defined boxes (also called anchors) set is $B$, each box $b_i \in B$ will be assigned a label $t_i \in \{-1,0,1,\ldots,K\}$ based on ground truth and the IoU strategy~\cite{girshick2015fast,ren2015faster}, where label $1\sim K$ means the object class ID, label ``0'' means background and label ``$-1$'' means ignored boxes. During training and testing phase, the detector outputs a score-vector $(s_i^0, \cdots, s_i^K)$ for each box $b_i$.

In our framework, instead of one box with $K+1$ dimensional score predictions,
we replicate each box $b_i$ for $K$  times to obtain $b_{ik}$ where $k=1,\cdots,K$, and the $k$-th box is responsible for the $k$-th class.
Each box $b_{ik}$ will be assigned a label $t_{ik} \in \{-1,0,1\}$ through the same IoU strategy (label $-1$ for not counted into the ranking loss).
Therefore, in the training and testing phase, the detector will predict only one scalar score $s_{ik}$ for each box $b_{ik}$.
\autoref{fig-labelassignment} illustrates our label formulation and the difference to traditional case.

The ranking task dictates that every positive boxes should be ranked higher than all the negative boxes w.r.t their scores.
Note that AP of our ranking result is computed over the scores from all classes.
This is slightly different from the evaluation metric meanAP for object detection systems, which computes AP for each class and obtains the average value.
We compute AP this way because the score distribution should be unified for all classes while ranking each class separately cannot achieve this goal.

\subsubsection{AP-Loss}
For simplicity, we still use $B$ to denote the anchor box set after replication, and $b_i$ to denote the $i$-th anchor box without the replication subscript. Each box $b_i$ thus corresponds to one scalar score $s_i$ and one binary label $t_i$.
Some transformations are required to formulate a ranking loss as illustrated in \autoref{fig-overall}.
First, the \textit{difference transformation} transfers the score $s_i$ to the difference form
\begin{equation}
\small
\forall i,j, \,\,\, x_{ij}=-(s(b_i; \bm{\theta})-s(b_j; \bm{\theta})) = -(s_i - s_j)
\end{equation}
where $s(b_i; \bm{\theta})$ is a CNN based score function with weights $\bm{\theta}$ for box \(b_i\).
The \textit{ranking label transformation} transfers labels $t_i$ to the corresponding pairwise ordering form
\begin{equation}
\small
\forall i,j, \,\,\, y_{ij} = \mathbf{1}_{t_i=1,t_j=0}
\end{equation}
where $\mathbf{1}$ is a indicator function which equals to 1 only if the subscript condition holds (\textit{i.e.}, $t_i=1,t_j=0$), otherwise $0$.
Then, we define an vector-valued \textit{activation function} $\bm{L}(\cdot)$ to produce the primary terms of the AP-loss as
\begin{equation}
\small
L_{ij}(\bm{x})=\frac{H(x_{ij})}{1+\sum_{k\in \mathcal{P}\cup\mathcal{N},k\neq i} H(x_{ik})} = L_{ij}
\label{Lij}
\end{equation}
where \(H(\cdot)\) is the Heaviside step function:
\begin{equation}
\small
H(x)=\left\{
\begin{aligned}
0& & x < 0 \\
1& & x\geq 0
\end{aligned}
\right.
\end{equation}

A ranking is denoted as \textit{proper ranking} when there are no two samples scored equally (\textit{i.e.}, $\forall i\neq j, \,\, s_i \neq s_j$). Without loss of generality, we will treat all rankings as a proper ranking by breaking ties arbitrarily. Now, we can formulate the AP-loss $\mathcal{L}_{AP}$ as
\begin{equation}
\small
\begin{split}
&\mathcal{L}_{AP} = 1- \text{AP} = 1-\frac{1}{|\mathcal{P}|}\sum_{i\in \mathcal{P}}\frac{rank^{+}(i)}{rank(i)} \\
& = 1-\frac{1}{|\mathcal{P}|} \sum_{i\in \mathcal{P}}\frac{1+\sum_{j\in \mathcal{P},j\neq i}H(x_{ij})}{1+\sum_{j\in \mathcal{P},j\neq i} H(x_{ij})+\sum_{j\in \mathcal{N}}H(x_{ij})} \\
& = \frac{1}{|\mathcal{P}|}\sum_{i\in \mathcal{P}}\sum_{j\in \mathcal{N}} L_{ij} = \frac{1}{|\mathcal{P}|} \sum_{i,j} L_{ij} \cdot y_{ij} = \frac{1}{|\mathcal{P}|} \langle \bm{L}(\bm{x}), \bm{y} \rangle
\end{split}
\label{Lap_1}
\end{equation}
where \(rank(i)\) and \(rank^{+}(i)\) denote the ranking position of score \(s_i\) among all valid samples and positive samples respectively, $\mathcal{P}=\{i | t_i=1\}$, $\mathcal{N}=\{i | t_i=0\}$, $|\mathcal{P}|$ is the size of set $\mathcal{P}$,
$\bm{L}$ and $\bm{y}$ are vector form for all $L_{ij}$ and $y_{ij}$ respectively, $\langle, \rangle$ means dot-product of two input vectors.
Note that $\bm{x},\bm{y}, \bm{L} \in \mathbb{R}^d$, where $d = (|\mathcal{P}|+|\mathcal{N}|)^2$.

Finally, the optimization problem can be written as:
\begin{equation}
\small
\min_{\bm{\theta}} \mathcal{L}_{AP}(\bm{\theta}) = 1-\text{AP}(\bm{\theta})=\frac{1}{|\mathcal{P}|} \langle \bm{L}(\bm{x}(\bm{\theta})), \bm{y}\rangle
\end{equation}
where $\bm{\theta}$ denotes the weights of detector model.
As the activation function $\bm{L}(\cdot)$ is non-differentiable, a novel optimization/learning scheme is required instead of the standard gradient descent method.

Besides the AP metric, other ranking based metric can also be used to design the ranking loss for our framework.
One example is the AUC-loss~\cite{li2013learning} which measures the area under ROC curve for ranking purpose, and has a slightly different ``activation function'' as
\begin{equation}
\small
L_{ij}'(\bm{x})=\frac{H(x_{ij})}{|\mathcal{N}|}
\end{equation}
As AP is consistent with the evaluation metric of the object detection task, we argue that AP-loss is intuitively more suitable than AUC-loss for this task,
and will provide empirical study in our experiments.

\subsection{Optimization Algorithm}
\subsubsection{Error-Driven Update}

Recalling the perceptron learning algorithm, the update for input variable is ``error-driven'', which means the update is directly derived from the difference between desired output and current output. We adopt this idea and further generalize it to accommodate the case of activation function with vector-valued input and output. Suppose $x_{ij}$ is the input and $L_{ij}$ is the current output, the update for $x_{ij}$ is thus
\begin{equation}
\small
\Delta x_{ij}=L^{*}_{ij}-L_{ij}
\label{error-driven}
\end{equation}
where \(L^{*}_{ij}\) is the desired output. Note that the AP-loss achieves its minimum possible value $0$ when each term $L_{ij}\cdot y_{ij} = 0$.
There are two cases. If $y_{ij}=1$, we should set the desired output $L^{*}_{ij}=0$.
If $y_{ij}=0$, we do not care the update and set it to $0$, since it does not contribute to the AP-loss.
Consequently, the update can be simplified as
\begin{equation}
\small
\Delta x_{ij}=-L_{ij} \cdot y_{ij}
\label{eq:update-xij}
\end{equation}

\subsubsection{Backpropagation}
We now have the desired vector-form update $\Delta \bm{x}$, and then will find an update for model weights $\Delta \bm{\theta}$  which will produce most appropriate movement for $\bm{x}$.
We use dot-product to measure the similarity of successive movements, and regularize the change of weights (\textit{i.e.} $\Delta\bm{\theta}$) with $L_2$-norm based penalty term. The optimization problem can be written as:
\begin{equation}
\small
\arg\min_{\Delta \bm{\theta}} \{-\langle \Delta \bm{x}, \bm{x}(\bm{\theta}^{(n)}+\Delta \bm{\theta})-\bm{x}(\bm{\theta}^{(n)}) \rangle +\lambda \| \Delta \bm{\theta} \|_2^2\}
\end{equation}
where $\bm{\theta}^{(n)}$ denotes the model weights at the $n$-th step. With that, the first-order expansion of \(\boldsymbol{x}(\boldsymbol{\theta})\) is given by:
\begin{equation}
\small
\bm{x}(\bm{\theta})=\bm{x}(\bm{\theta}^{(n)})+\frac{\partial \bm{x}(\bm{\theta}^{(n)})}{\partial \bm{\theta}} \cdot (\bm{\theta}-\bm{\theta}^{(n)})
 + o(\|\bm{\theta}-\bm{\theta}^{(n)}\|)
\end{equation}
where $\partial \bm{x}(\bm{\theta}^{(n)})/ \partial \bm{\theta}$ is the Jacobian matrix of vector-valued function $\bm{x}(\bm{\theta})$ at $\bm{\theta}^{(n)}$.
Ignoring the high-order infinitesimal, we obtain the step-wise minimization process:
\begin{equation}
\small
\bm{\theta}^{(n+1)}-\bm{\theta}^{(n)}=\arg\min_{\Delta\bm{\theta}}\{-\langle \Delta \bm{x}, \frac{\partial \bm{x}(\bm{\theta}^{(n)})}{\partial \bm{\theta}} \Delta\bm{\theta} \rangle  + \lambda \| \Delta\bm{\theta}\|_2^2\}
\end{equation}
The optimal solution can be obtained by finding the stationary point. Then, the form of optimal \(\Delta\bm{\theta}\) is consistent with the chain rule of derivative, which means, it can be directly implemented by setting the gradient of \(x_{ij}\) to \(-\Delta x_{ij}\) (c.f. \autoref{eq:update-xij}) and proceeding with backpropagation. Hence the gradient for score \(s_i\) can be obtained by backward propagating the gradient through the difference transformation:
\begin{equation}
\small
\begin{split}
g_i= - \sum_{j,k} & \Delta x_{jk} \cdot \frac{\partial x_{jk}}{\partial s_i} = \sum_j \Delta x_{ij} - \sum_j \Delta x_{ji} \\
& = \sum_j L_{ji} \cdot y_{ji} - \sum_j L_{ij} \cdot y_{ij}
\end{split}
\label{gradient}
\end{equation}

\subsection{Analyses}
{\bf \noindent Convergence:}
To better understand the characteristics of the AP-loss, we first provide a theoretical analysis on the convergence of the optimization algorithm, which is generalized from the convergence property of the original perceptron learning algorithm.
\begin{proposition}
The AP-loss optimizing algorithm is guaranteed to converge in finite steps if below conditions hold:
{\noindent (1) the learning model is linear;} \par
{\noindent (2) the training data is linearly separable.}
\end{proposition}
The proof of this proposition is provided in Appendix-1 of supplementary.
Although convergence is somewhat weak due to the need of strong conditions, it is \textbf{non-trivial} since the AP-loss function is not convex or quasiconvex even for the case of linear model and linearly separable data, so that gradient descent based algorithm may still fail to converge on a smoothed AP-loss function even under such strong conditions. One such example is presented in Appendix-2 of supplementary. It means that, under such conditions, our algorithm still optimizes better than the approximate gradient descent algorithm for AP-loss.
Furthermore, with some mild modifications, even though the training data is not separable, the accumulated AP-loss can also be bounded proportionally by the best performance of the learning model. More details are presented in Appendix-3 of supplementary.

{\bf \noindent Consistency:}
Besides convergence, We observed that the proposed optimization algorithm is inherently consistent with widely used classification-loss functions.
\begin{observation}
When the activation function \(L(\cdot)\) takes the form of softmax function and loss-augmented step function, our optimization algorithm can be expressed as the gradient descent algorithm on cross-entropy loss and hinge loss respectively.
\label{consistency}
\end{observation}
The detailed analysis of this observation is presented in Appendix-4 of supplementary. We argue that the observed consistency is on the basis of the ``error-driven'' property.
As is known, the gradients of those widely used loss functions are proportional to their prediction errors, where the prediction here refers to the output of activation function. In other words, their activation functions have a nice property: the vector field of prediction errors is conservative, allowing it being the gradient of some surrogate loss function.
However, our activation function does not have this property, which makes our optimization not able to express as gradient descent with any surrogate loss function.

\subsection{Details of Training Algorithm} \label{training-detail}

\begin{algorithm}[t]
\footnotesize
\caption{Minibatch training for Interpolated AP}
\begin{algorithmic}[1]
  \Require All scores \(\{s_i\}\) and corresponding labels \(\{t_i\}\) in a minibatch
  \Ensure Gradient of input \(\{g_i\}\)
  \State $\forall i, \,\,\, g_i \gets 0$
  \State $\text{MaxPrec} \gets 0$
  \State $\mathcal{P} \gets \{i \mid t_i=1\}, \,\,\, \mathcal{N} \gets \{i \mid t_i=0\}$
  \State $O \gets argsort(\{s_i \mid i\in \mathcal{P}\}) $ \Comment{Indexes of scores sorted in ascending order}
  \For {$i \in O$}
    \State Compute \(x_{ij}=s_j-s_i\) for all \(j\in \mathcal{P}\cup\mathcal{N}\) and \(L_{ij}\) for all \(j\in \mathcal{N}\) \Comment{According to \autoref{Lij} and \autoref{smooth}}
    \State $\text{Prec} \gets 1-\sum_{j \in \mathcal{N}} L_{ij}$
    \If {$\text{Prec} \geq \text{MaxPrec}$}
      \State $ \text{MaxPrec} \gets \text{Prec}$
    \Else \Comment{Interpolation}
      \State $\forall j \in \mathcal{N}, \,\,\, L_{ij} \gets L_{ij} \cdot (1-\text{MaxPrec}) / (1-\text{Prec})$
    \EndIf
    \State $g_i \gets -\sum_{j\in \mathcal{N}} L_{ij}$ \Comment{According to \autoref{gradient}}
    \State $\forall j \in \mathcal{N}, \,\,\, g_j \gets g_j+L_{ij}$ \Comment{According to \autoref{gradient}}
  \EndFor
  \State $\forall i, \,\,\, g_i \gets g_i / |\mathcal{P}|$ \Comment{Normalization}
\end{algorithmic}
\label{IAP}
\end{algorithm}

\begin{table*}[t]
\small
\centering
\begin{subtable}[b]{0.3\linewidth}
\centering
\begin{tabular}{c|ccc}
\hline
Batch Size & AP & AP\(_{50}\) & AP\(_{75}\) \\
\hline\hline
1 & 52.4 & 80.2 & 56.7 \\
2 & 53.0 & 81.7 & 57.8 \\
4 & 52.8 & 82.2 & 58.0 \\
8 & \textbf{53.1} & \textbf{82.3} & \textbf{58.1} \\
\hline
\end{tabular}
\caption{Varying batch size}
\label{voc-minibatch-training}
\end{subtable}
\quad
\begin{subtable}[b]{0.3\linewidth}
\centering
\begin{tabular}{c|ccc}
\hline
\(\delta\) & AP & AP\(_{50}\) & AP\(_{75}\) \\
\hline\hline
0.25 & 50.2 & 80.7 & 53.6 \\
0.5 & 51.3 & 81.6 & 55.4 \\
1 & \textbf{53.1} & 82.3 & \textbf{58.1} \\
2 & 52.8 & \textbf{82.6} & 57.2 \\
\hline
\end{tabular}
\caption{Varying \(\delta\) for piecewise step function}
\label{voc-smoothing}
\end{subtable}
\quad
\begin{subtable}[b]{0.3\linewidth}
\centering
\begin{tabular}{c|ccc}
\hline
Interpolated  & AP & AP\(_{50}\) & AP\(_{75}\) \\
\hline\hline
No & 52.6 & 82.2 & 57.1 \\
Yes & \textbf{53.1} & \textbf{82.3} & \textbf{58.1} \\
\hline
\end{tabular}
\caption{Interpolated \textit{vs.} not interpolated}
\label{voc-interpolated-ap}
\end{subtable}
\vspace{-3mm}
\caption{Ablation experiments. Models are tested on VOC2007 {\tt test} set. }
\vspace{-5mm}
\label{voc-abaltion-study}
\end{table*}

{\bf\noindent Minibatch Training}
The minibatch training strategy is widely used in deep learning frameworks~\cite{he2016deep,liu2016ssd,lin2018focal} as it accounts for more stability than the case with batch size equal to 1.
The mini-batch training helps our optimization algorithm quite a lot for escaping the so-called ``score-shift'' scenario.
The AP-loss can be computed both from a batch of images and from a single image with multiple anchor boxes.
Consider an extreme case: our detector can predict perfect ranking in both image $I_1$ and image $I_2$, but the lowest score in image $I_1$ is even greater than the highest score in image $I_2$. There are ``score-shift'' between two images so that the detection performance is poor when computing AP-loss per-image.
Aggregating scores over images in a mini-batch can avoid such problem, so that the minibatch training is crucial for good convergence and good performance.

\begin{figure}[t]
\centering
  \includegraphics[width=0.9\linewidth]{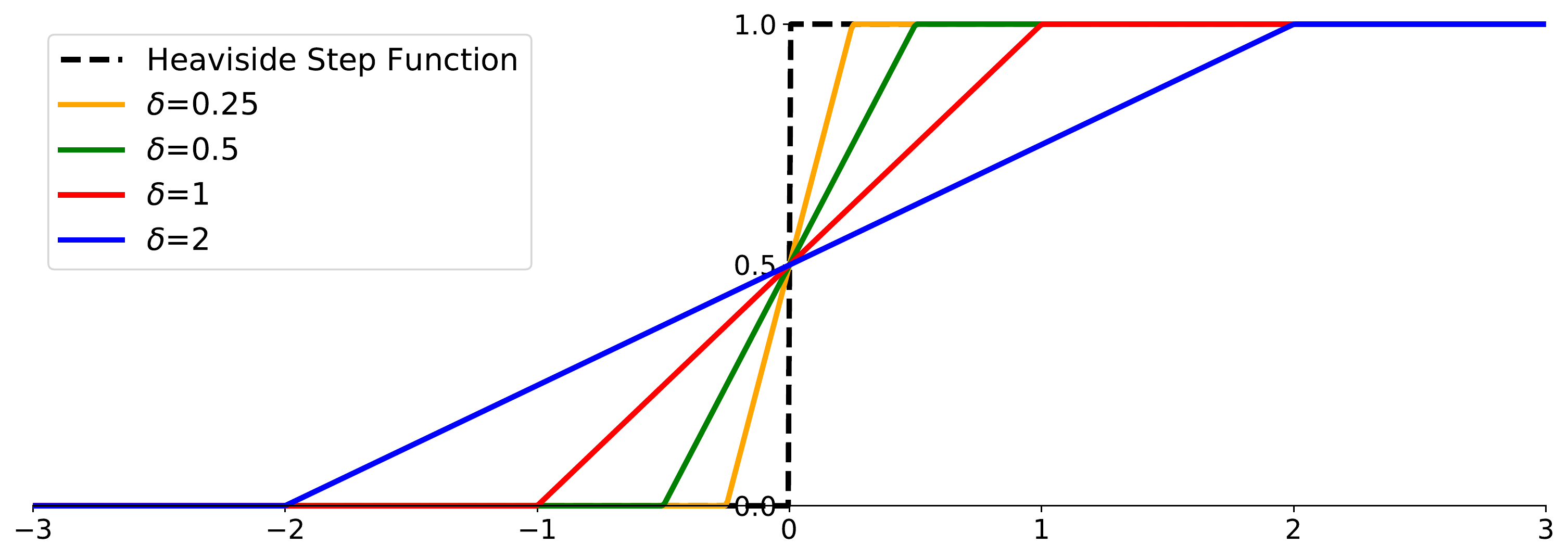}
\vspace{-3mm}
  \caption{Heaviside step function and piecewise step function. (Best viewed in color)}
\vspace{-5mm}
\label{fig-step_function}
\end{figure}

{\bf\noindent Piecewise Step function}
During early stage of training, scores $s_i$ are very close to each other (\textit{i.e.} almost all inputs to Heaviside step function $H(x)$ are near zero), so that a small change of input will cause a big output difference, which destabilizes the updating process. To tackle this issue, we replace $H(x)$ with a piecewise step function:
\begin{equation}
\small
f(x)=\left\{
\begin{aligned}
&0\, , & x < -\delta \\
&\frac{x}{2\delta}+0.5\, , & -\delta \leq x \leq \delta \\
&1\, , & \delta < x
\end{aligned}
\right.
\label{smooth}
\end{equation}
The piecewise step functions with different \(\delta\) are shown in \autoref{fig-step_function}. When \(\delta\) approaches \(+0\), the piecewise step function approaches the original step function. Note that \(f(\cdot )\) is only different from \(H(\cdot )\) near zero point. We argue that the precise form of the piecewise step function is not crucial. Other monotonic and symmetric smooth functions that only differs from \(H(\cdot )\) near zero point could be equally effective. The choice of $\delta$ relates closely to the weight decay hyper-parameter in CNN optimization. Intuitively, parameter $\delta$ controls the width of decision boundary between positive and negative samples. Smaller $\delta$ enforces a narrower decision boundary, which causes the weights to shrink correspondingly (similar effect to that caused by the weight decay). Further details are presented in the experiments.

{\bf\noindent Interpolated AP}
The interpolated AP~\cite{salton1986introduction} is widely adopted by many object detection benchmarks like PASCAL VOC~\cite{everingham2015pascal} and MS COCO~\cite{lin2014microsoft}.
The common justification for interpolating the precision-recall curve~\cite{everingham2015pascal} is ``to reduce the impact of 'wiggles' in the precision-recall curve, caused by small variations in the ranking of examples''.
Under the same consideration, we adopt the interpolated AP instead of the original version. Specifically, the interpolation is applied on $L_{ij}$ to make the precision at the $k$-th smallest positive sample monotonically increasing with \(k\) where the precision is $(1-\sum_{j\in \mathcal{N}} L_{ij})$ in which $i$ is the index of the $k$-th smallest positive sample. It is worth noting that the interpolated AP is a smooth approximation of the actual AP so that it is a practical choice to help to stabilize the gradient and to reduce the impact of 'wiggles' in the update signals. The details of the interpolated AP based algorithm is summarized in Algorithm \ref{IAP}.

\section{Experiments}
\subsection{Experimental Settings}
We evaluate the proposed method on the state-of-the-art one-stage detector RetinaNet~\cite{lin2018focal}. The implementation details are the same as in~\cite{lin2018focal} unless explicitly stated. Our experiments are performed on two benchmark datasets: PASCAL VOC~\cite{everingham2015pascal} and MS COCO~\cite{lin2014microsoft}. The PASCAL VOC dataset has 20 classes, with VOC2007 containing 9,963 images for train/val/test and VOC2012 containing 11,530 for train/val. The MS COCO dataset has 80 classes, containing 123,287 images for train/val. We implement our codes with the MXNET framework, and conduct experiments on a workstation with two NVidia TitanX GPUs.

{\bf \noindent PASCAL VOC:} When evaluated on the VOC2007 {\tt test} set, models are trained on the VOC2007 and VOC2012 {\tt trainval} sets. When evaluated on the VOC2012 {\tt test} set, models are trained on the VOC2007 and VOC2012 {\tt trainval} sets plus the VOC2007 {\tt test} set. Similar to the evaluation metrics used in the MS COCO benchmark, we also report the AP averaged over multiple IoU thresholds of \(0.50:0.05:0.95\). We set \(\delta=1\) in \autoref{smooth}. We use ResNet~\cite{he2016deep} as the backbone model which is pre-trained on the ImageNet-1k classification dataset~\cite{deng2009imagenet}. At each level of FPN~\cite{lin2017feature}, the anchors have 2 sub-octave scales (\(2^{k/2}\), for \(k\leq 1\)) and 3 aspect ratios [0.5, 1, 2]. We fix the batch normalization layers to be frozen in training phase. We adopt the minibatch training on 2 GPUs with 8 images per GPU. All evaluated models are trained for 160 epochs with an initial learning rate of 0.001 which is then divided by 10 at 110 epochs and again at 140 epochs. Weight decay of 0.0001 and momentum of 0.9 are used. We adopt the same data augmentation strategies as~\cite{liu2016ssd}, while do not use any data augmentation during testing phase. In training phase, the input image is fixed to 512\(\times\)512, while in testing phase, we maintain the original aspect ratio and resize the image to ensure the shorter side with 600 pixels. We apply the non-maximum suppression with IoU of 0.5 for each class.

{\bf \noindent MS COCO:} All models are trained on the widely used {\tt trainval35k} set (80k train images and 35k subset of val images), and tested on {\tt minival} set (5k subset of val images) or {\tt test-dev} set. We train the networks for 100 epochs with an initial learning rate of 0.001 which is then divided by 10 at 60 epochs and again at 80 epochs. Other details are similar to that for PASCAL VOC.

\subsection{Ablation Study}
We first investigate the impact of our design settings of the proposed framework. We fix the ResNet-50 as backbone and conduct several controlled experiments on PASCAL VOC2007 {\tt test} set (and COCO {\tt minival} if stated) for this ablation study.

\subsubsection{Comparison on Different Parameter Settings}

\begin{table}[t]
\small
\centering
\setlength{\tabcolsep}{1.3mm}{
\begin{tabular}{c|ccc|ccc}
\hline
\multirow{2}*{Training Loss} & \multicolumn{3}{c|}{PASCAL VOC} & \multicolumn{3}{c}{COCO} \\
\cline{2-7}
 & AP & AP\(_{50}\) & AP\(_{75}\) & AP & AP\(_{50}\) & AP\(_{75}\)\\
\hline\hline
CE-Loss + OHEM & 49.1 & 81.5 & 51.5 & 30.8 & 50.9 & 32.6\\
Focal Loss & 51.3 & 80.9 & 55.3 & 33.9 & 55.0 & 35.7 \\
\cline{1-7}
AUC-Loss & 49.3 & 79.7 & 51.8 & 25.5 & 44.9 & 26.0 \\
AP-Loss & \textbf{53.1} & \textbf{82.3} & \textbf{58.1} & \textbf{35.0} & \textbf{57.2} & \textbf{36.6}\\
\hline
\end{tabular}}
\vspace{-2mm}
\caption{Comparison through different training losses. Models are tested on VOC2007 {\tt test} and COCO {\tt minival} sets. The metric AP is averaged over multiple IoU thresholds of \(0.50:0.05:0.95\).}
\vspace{-3mm}
\label{base-loss}
\end{table}

Here we study the impact of the practical modifications introduced in Section \ref{training-detail}. All results are shown in \autoref{voc-abaltion-study}.

{\bf \noindent Minibatch Training:}
First, we study the mini-batch training, and report detector results at different batch-size in \autoref{voc-minibatch-training}. It shows that larger batch-size (\textit{i.e.} 8) outperforms all the other smaller batch-size. This verifies our previous hypothesis that large minibatch training helps to eliminate the ``score-shift'' from different images, and thus stabilizes the AP-loss through robust gradient calculation. Hence, \(\text{batch-size}=8\) is used in our further studies.

{\bf \noindent Piecewise Step Function:}
Second, we study the piecewise step function, and report detector performance on the piecewise step function with different \(\delta\) in \autoref{voc-smoothing}. As mentioned before, we argue that the choice of \(\delta\) is trivial and is dependent on other network hyper-parameters such as weight decay. Smaller \(\delta\) makes the function sharper, which yields unstable training at initial phase. Larger \(\delta\) makes the function deviate from the properties of the original AP-loss, which also worsens the performance. \(\delta=1\) is a good choice we used in our further studies.

{\bf \noindent Interpolated AP:}
Third, we study the impact of interpolated AP in our optimization algorithm, and list the results in \autoref{voc-interpolated-ap}. Marginal benefits are observed for interpolated AP over standard AP, so we use interpolated AP in all the following studies.

\subsubsection{Comparison on Different Losses}

\begin{figure}[t]
\centering
\small
\begin{subfigure}[b]{0.49\linewidth}
  \centering
  \includegraphics[width=1.0\linewidth]{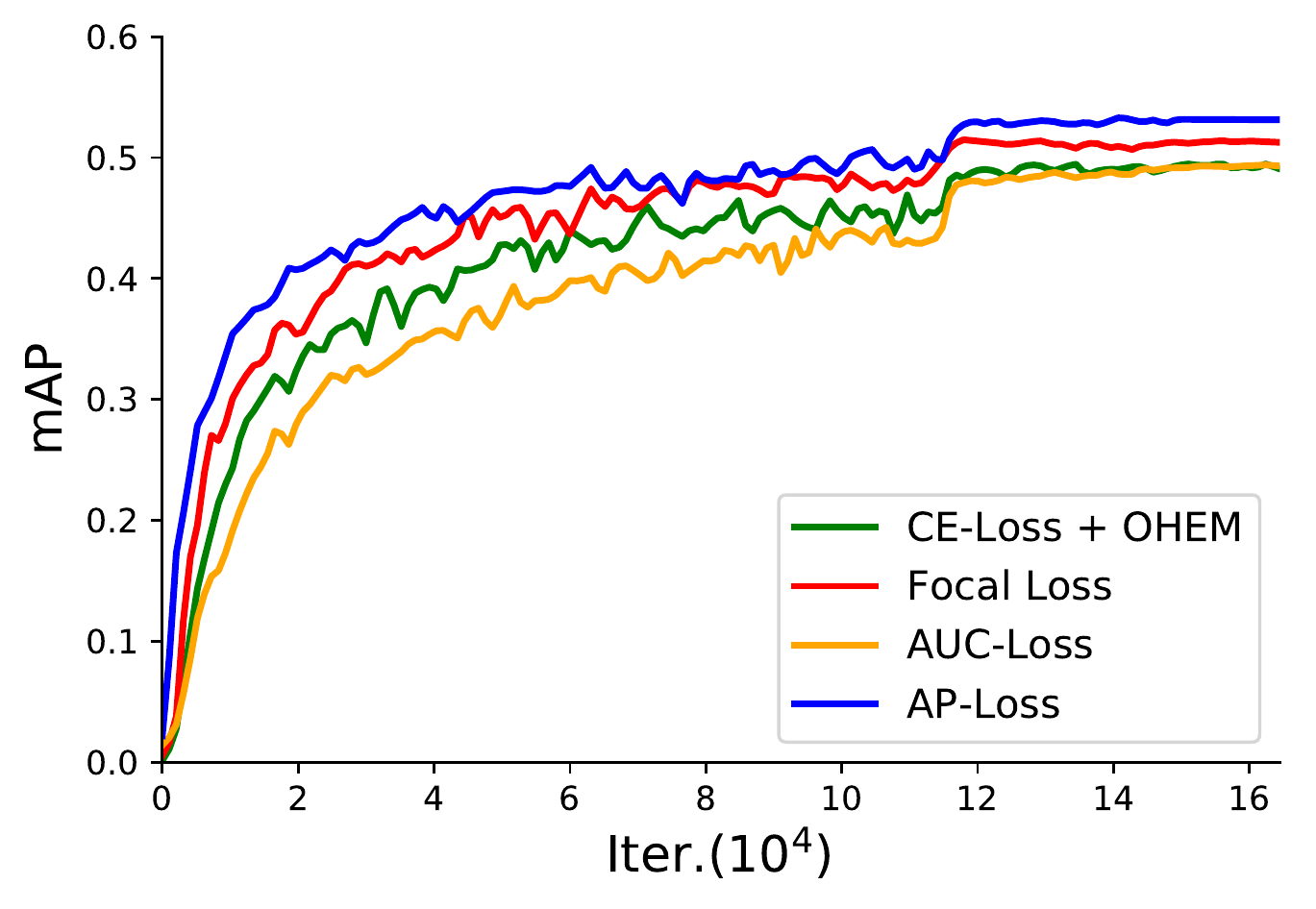}
  \vspace{-6mm}
  \caption{}
  \label{fig-performance_curve}
\end{subfigure}
\begin{subfigure}[b]{0.49\linewidth}
  \centering
  \includegraphics[width=1.0\linewidth]{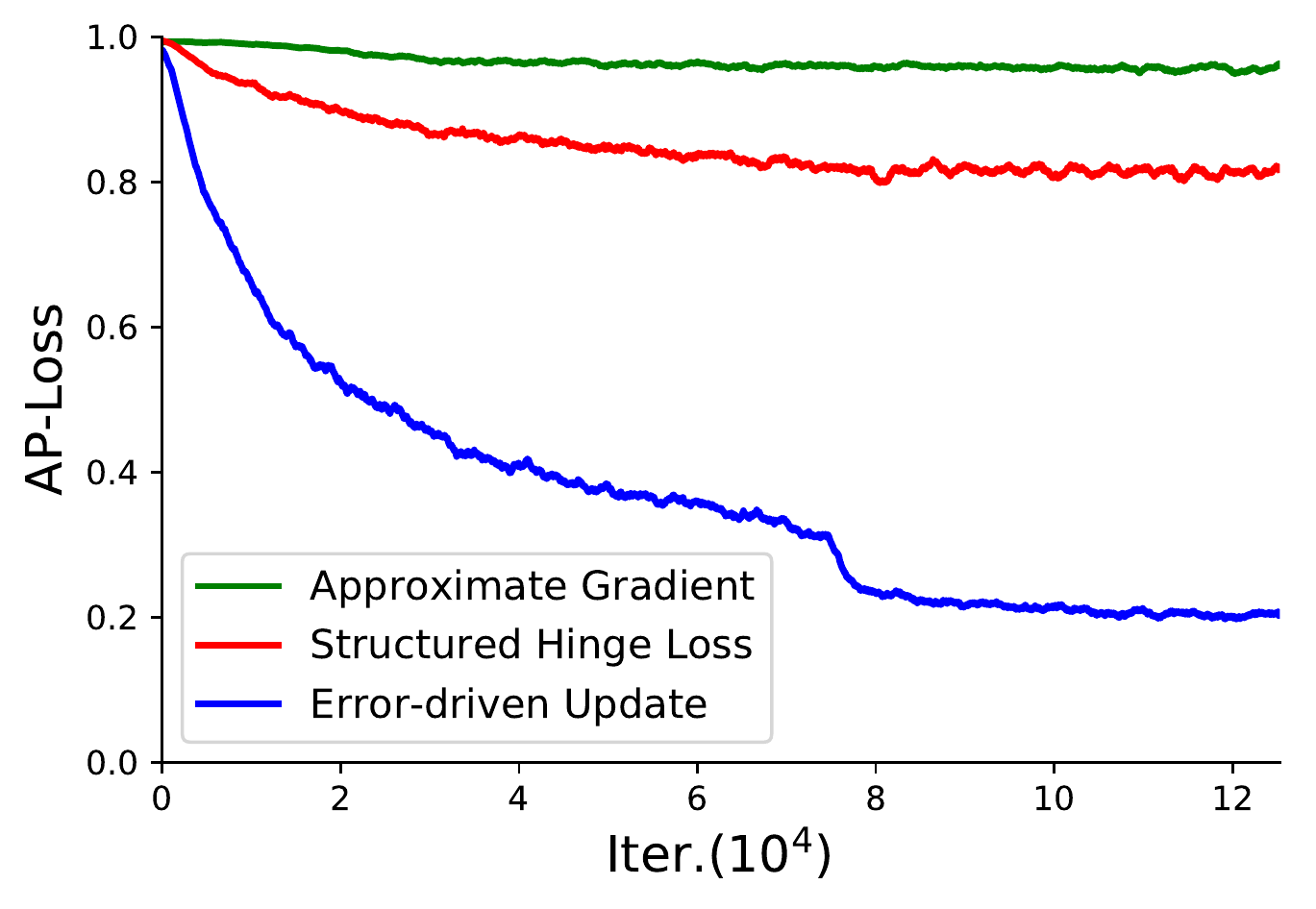}
  \vspace{-6mm}
  \caption{}
  \label{fig-convergence_curve}
\end{subfigure}
\vspace{-3mm}
  \caption{(a) Detection accuracy (mAP) on VOC2007 {\tt test} set. (b) Convergence curves of different AP-loss optimizations on VOC2007 {\tt trainval} set. (Best viewed in color)}
\vspace{-3mm}
\end{figure}

We evaluate with different losses on RetinaNet~\cite{lin2018focal}. Results are shown in \autoref{base-loss}. We compare traditional classification based losses like focal loss~\cite{lin2018focal} and cross entropy loss (CE-loss) with OHEM~\cite{liu2016ssd} to the ranking based losses like AUC-loss and AP-loss. Although focal loss is significantly better than CE-loss with OHEM on COCO dataset, it is interesting that focal-loss does not perform better than CE-loss at AP\(_{50}\) on PASCAL VOC. This is likely because the hyper-parameters of focal loss are designed to suit the imbalance condition on COCO dataset which is not suitable for PASCAL VOC, so that focal loss cannot generalize well to PASCAL VOC without tuning its hyper-parameters. The proposed AP-loss performs much better than all the other losses on both two datasets, which demonstrates its effectiveness and stronger generalization ability on handling the imbalance issue. It is worth noting that AUC-loss performs much worse than AP-loss, which may be due to the fact that AUC has equal penalty for each misordered pair while AP imposes greater penalty for the misordering at higher positions in the predicted ranking. It is obvious that object detection evaluation concerns more on objects with higher confidence, which is why AP provides a better loss measure. Furthermore, an assessment of the detection performance at different training iterations, as shown in \autoref{fig-performance_curve}, outlines the superiority of the AP-loss for snapshot time points.

\begin{figure*}[t!]
\small
\begin{subfigure}[b]{0.188\linewidth}
  \centering
  \includegraphics[width=1.0\linewidth]{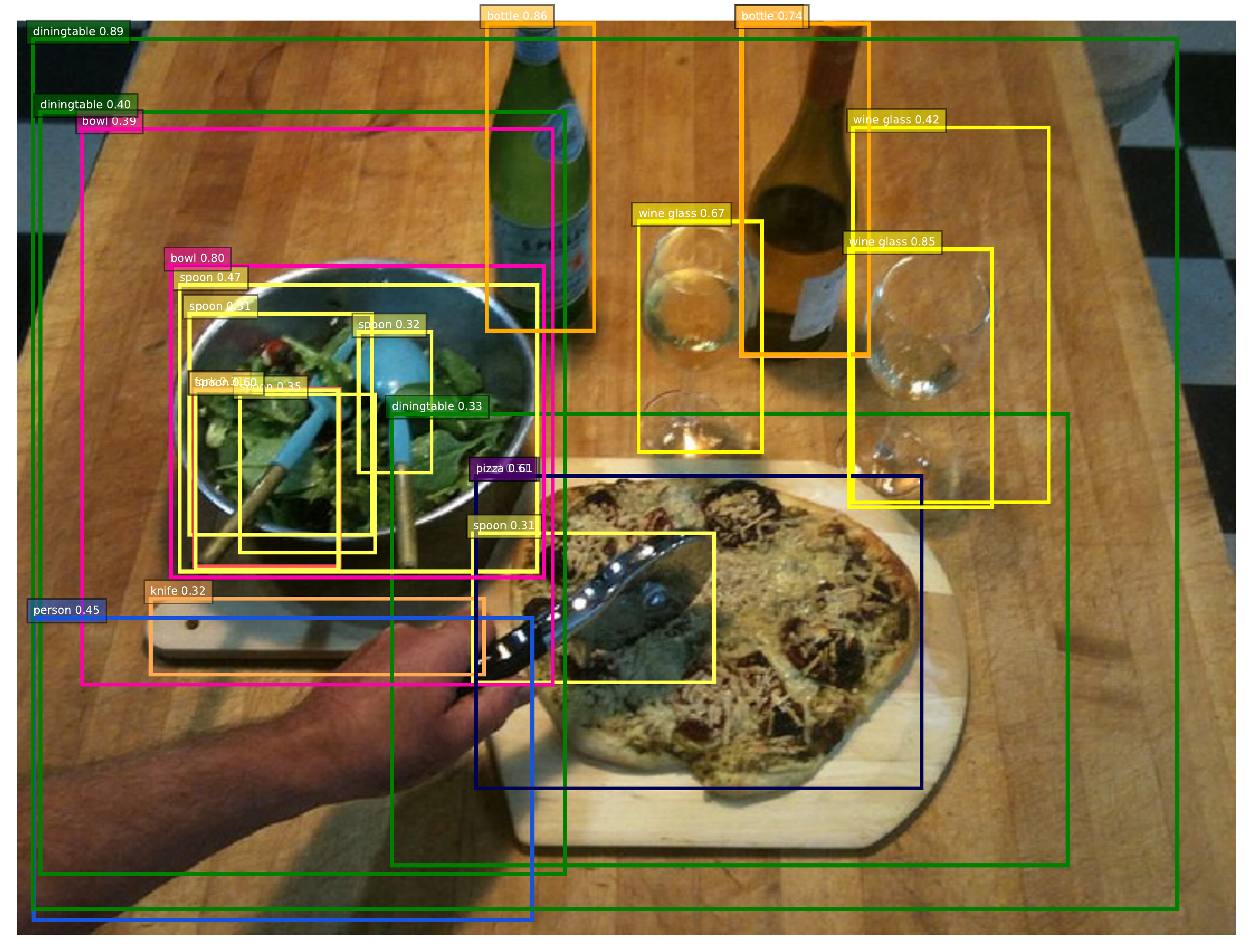}
\end{subfigure}
\begin{subfigure}[b]{0.21\linewidth}
  \centering
  \includegraphics[width=1.0\linewidth]{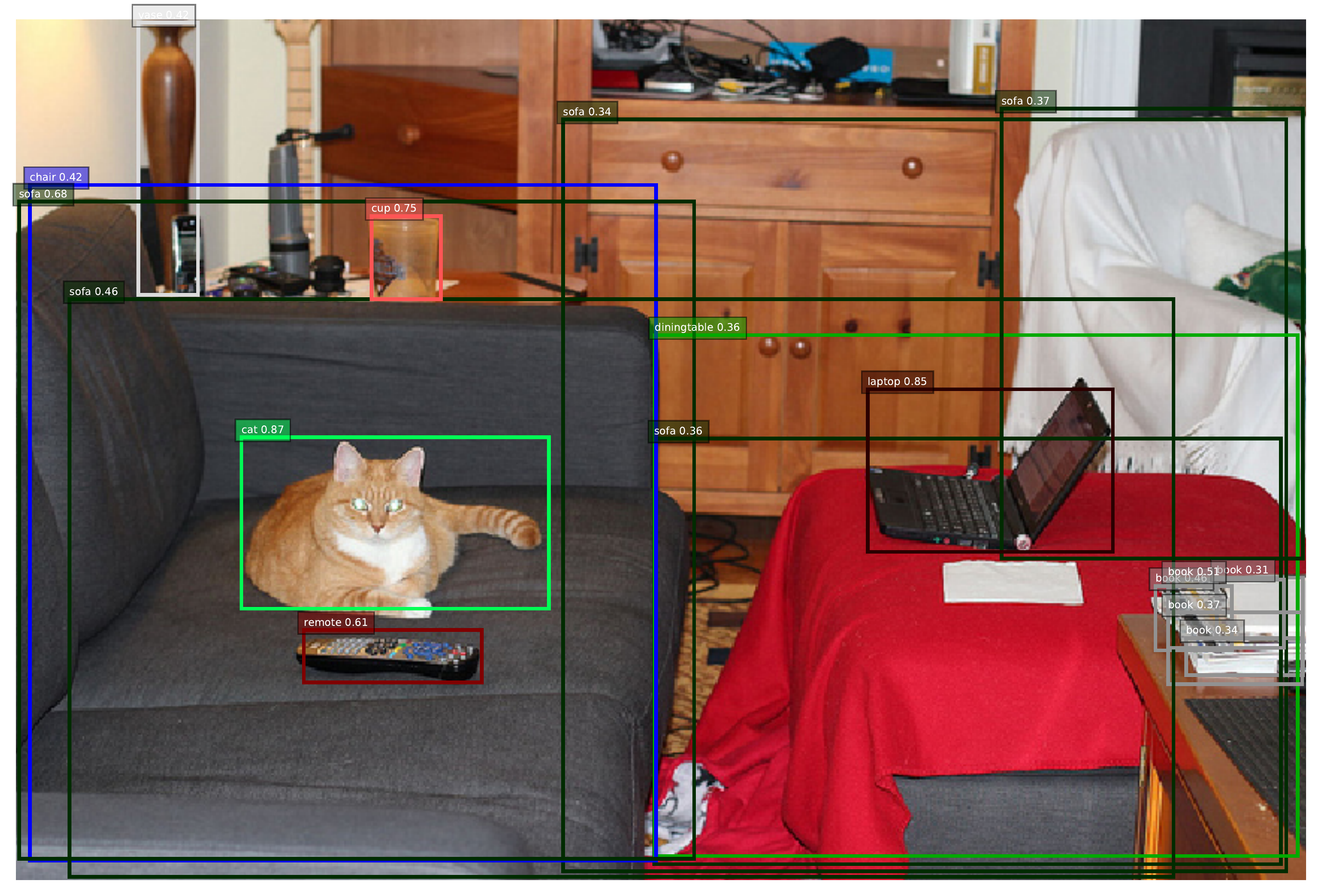}
\end{subfigure}
\begin{subfigure}[b]{0.21\linewidth}
  \centering
  \includegraphics[width=1.0\linewidth]{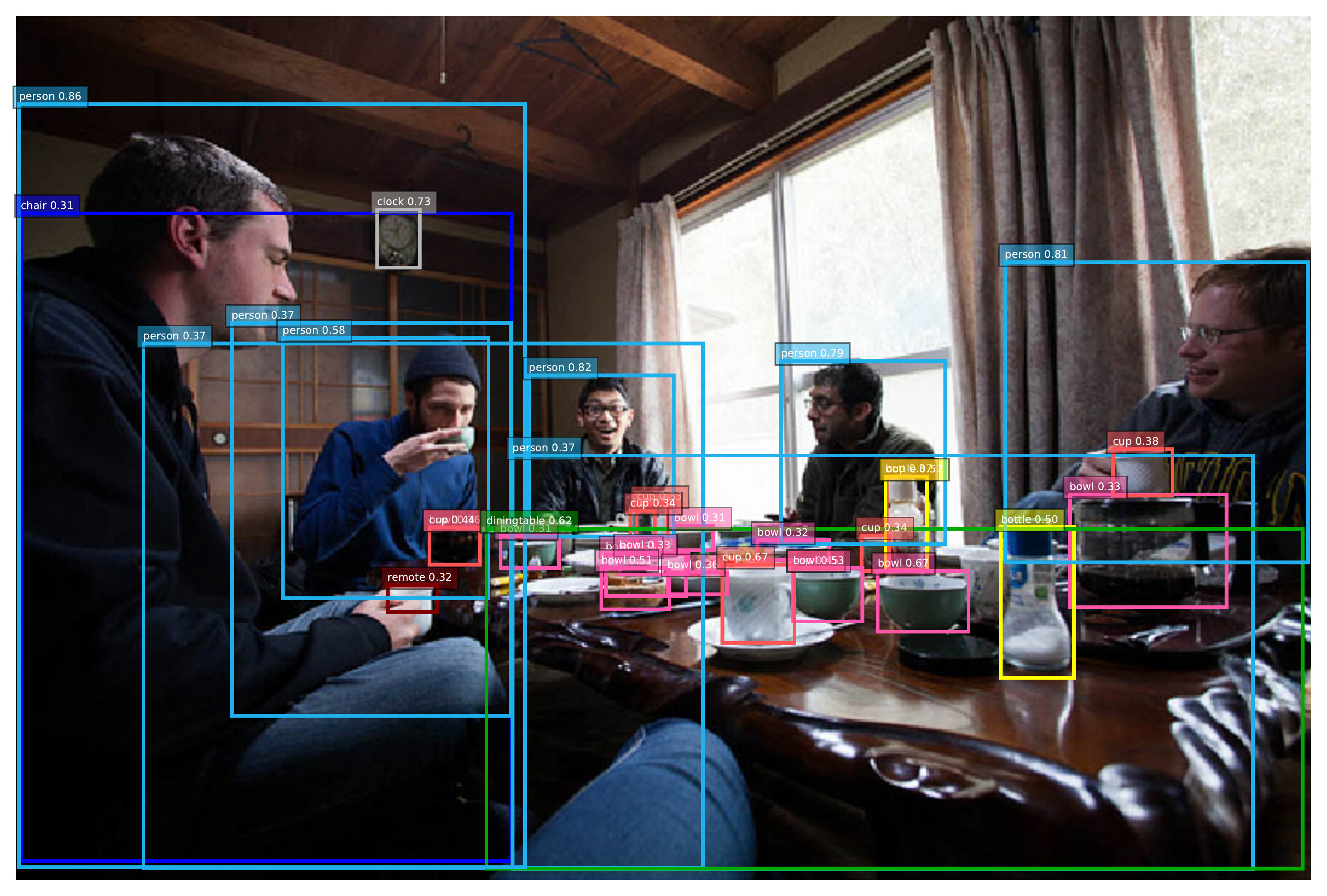}
\end{subfigure}
\begin{subfigure}[b]{0.188\linewidth}
  \centering
  \includegraphics[width=1.0\linewidth]{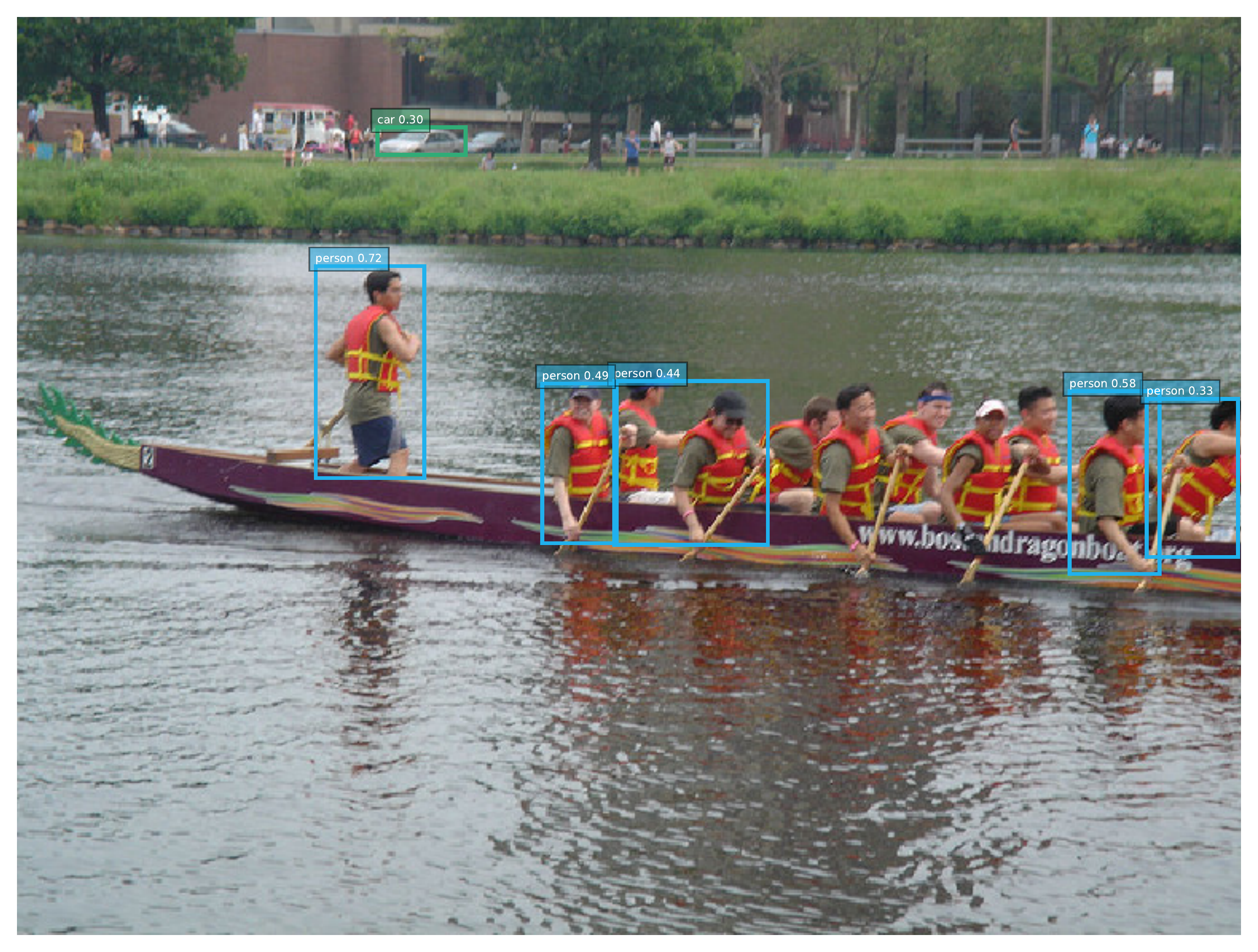}
\end{subfigure}
\begin{subfigure}[b]{0.188\linewidth}
  \centering
  \includegraphics[width=1.0\linewidth]{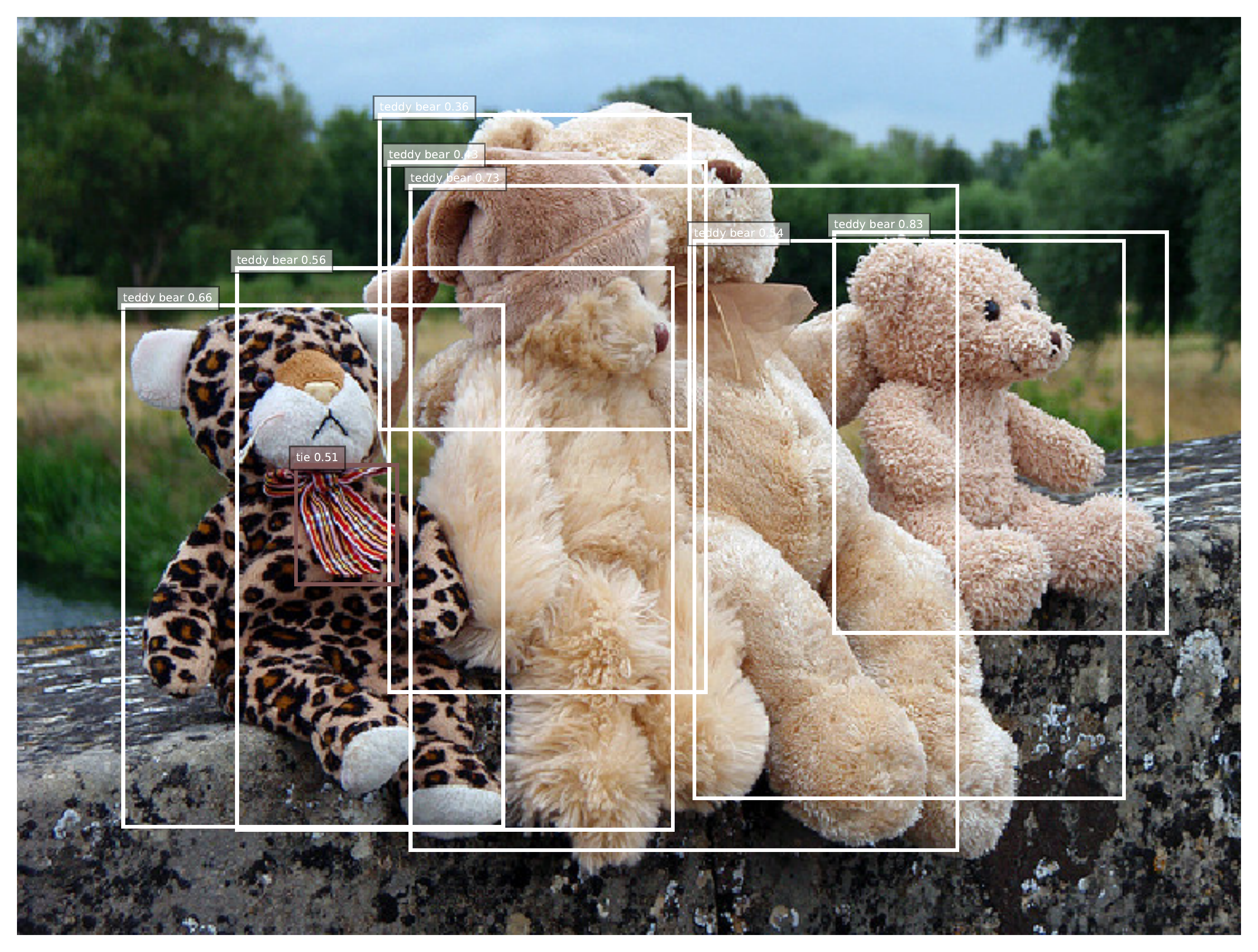}
\end{subfigure}
\par
\begin{subfigure}[b]{0.188\linewidth}
  \centering
  \includegraphics[width=1.0\linewidth]{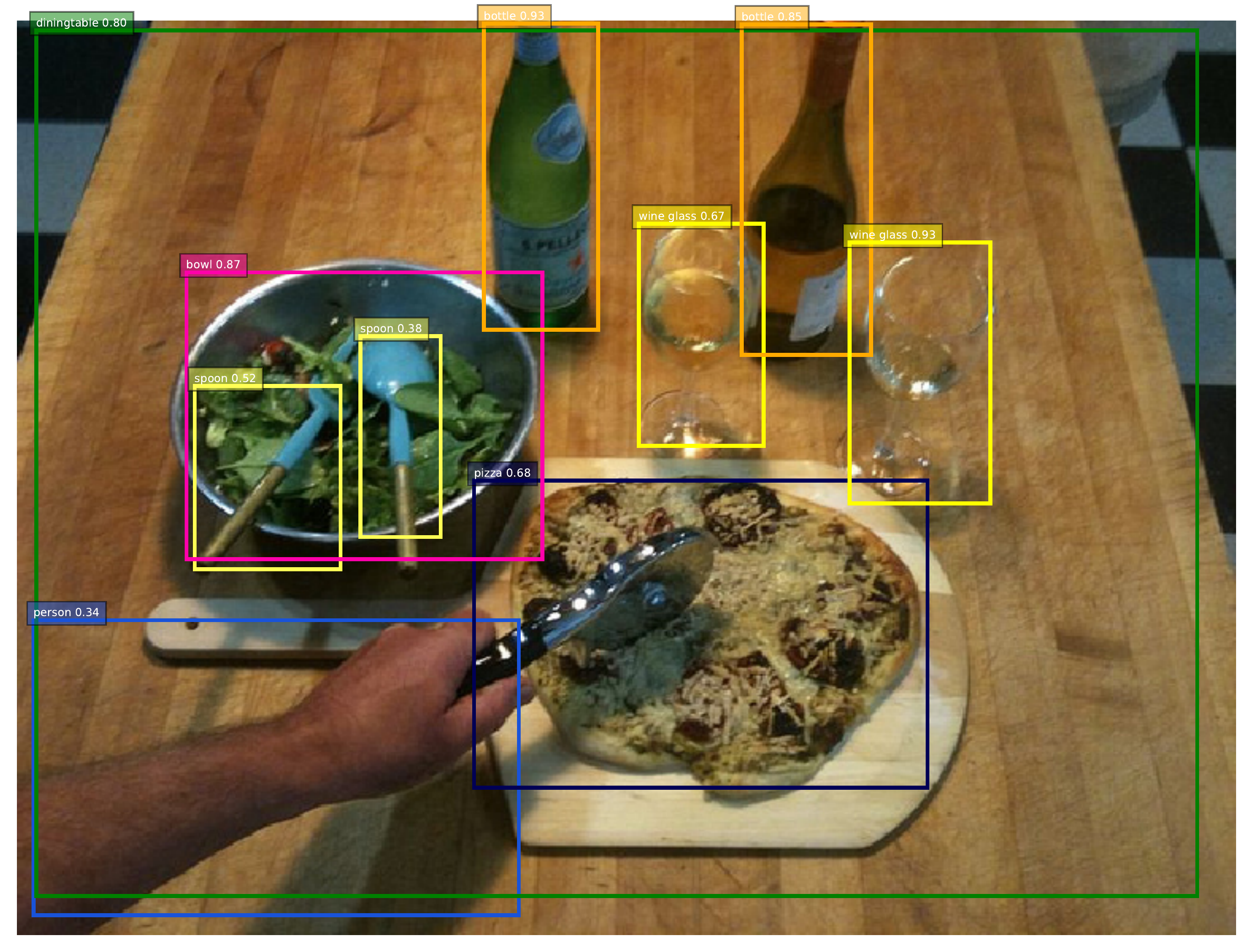}
\end{subfigure}
\begin{subfigure}[b]{0.21\linewidth}
  \centering
  \includegraphics[width=1.0\linewidth]{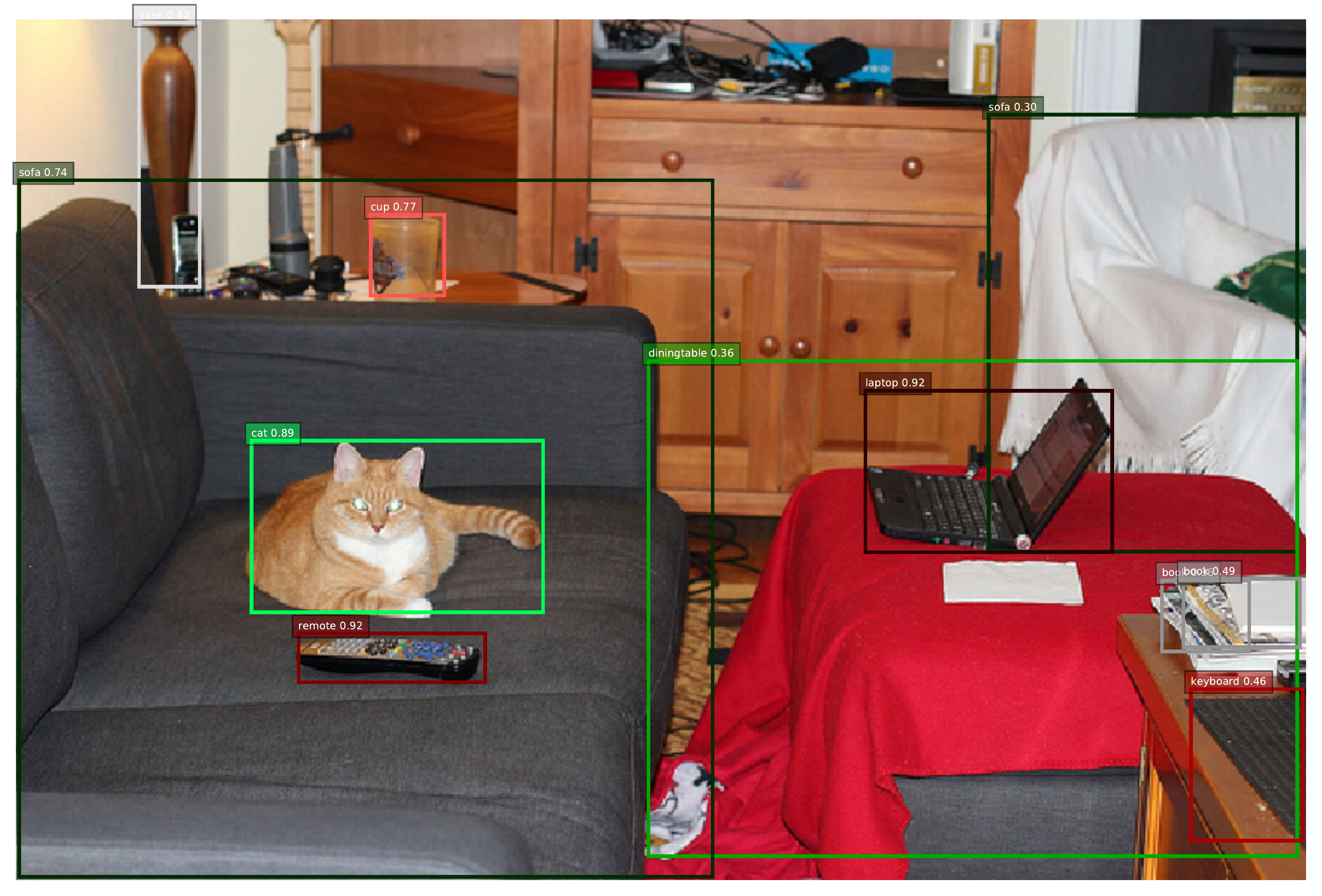}
\end{subfigure}
\begin{subfigure}[b]{0.21\linewidth}
  \centering
  \includegraphics[width=1.0\linewidth]{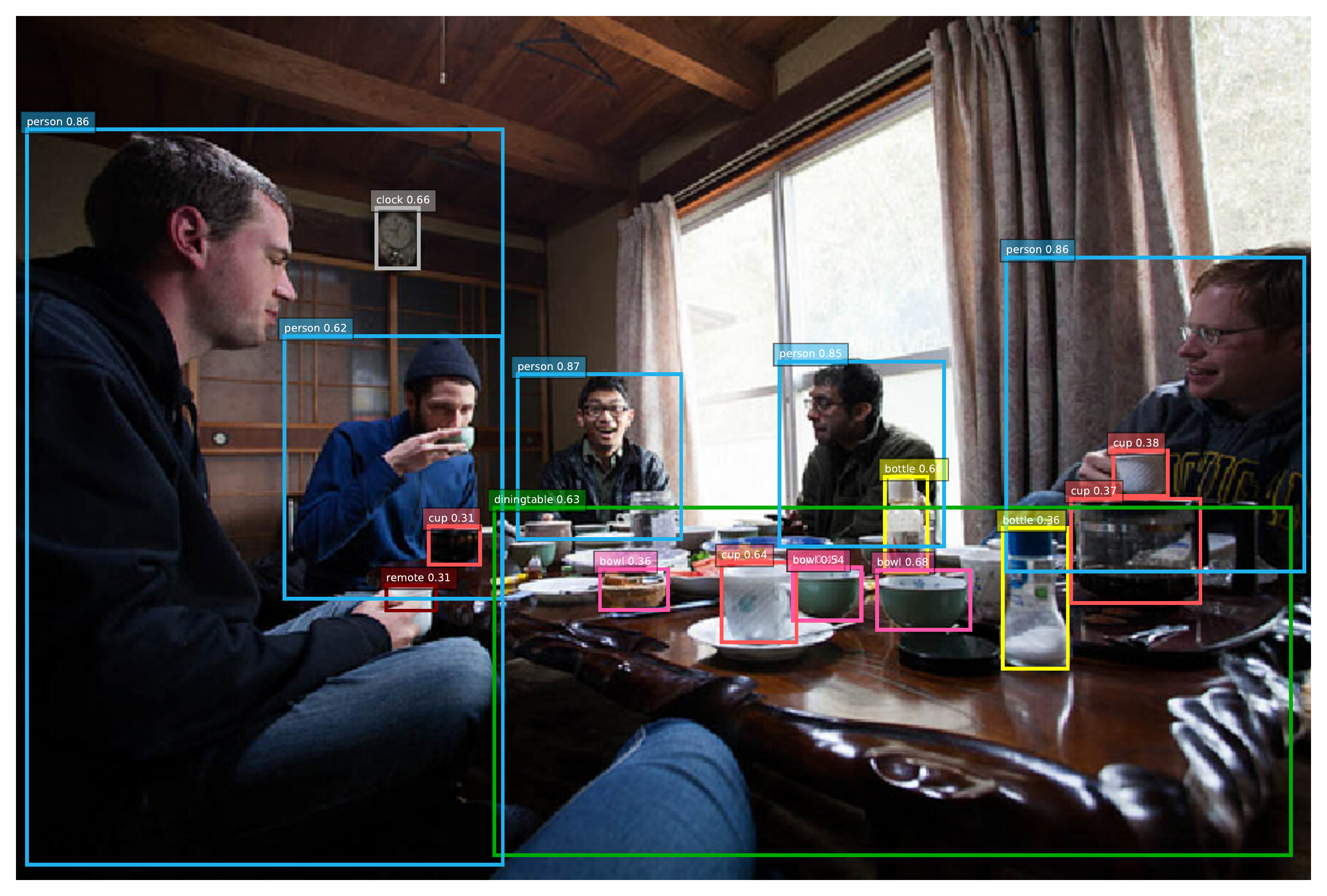}
\end{subfigure}
\begin{subfigure}[b]{0.188\linewidth}
  \centering
  \includegraphics[width=1.0\linewidth]{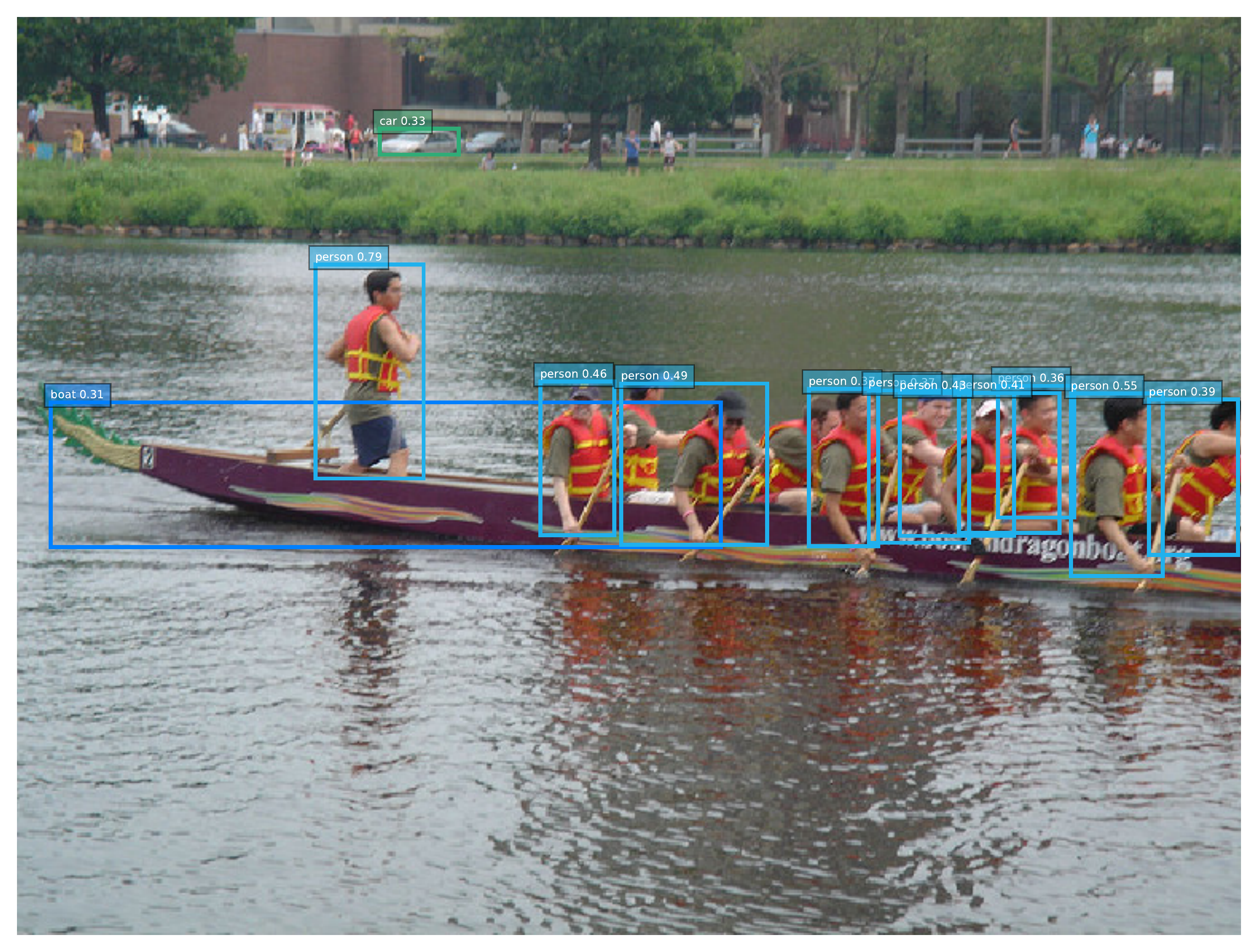}
\end{subfigure}
\begin{subfigure}[b]{0.188\linewidth}
  \centering
  \includegraphics[width=1.0\linewidth]{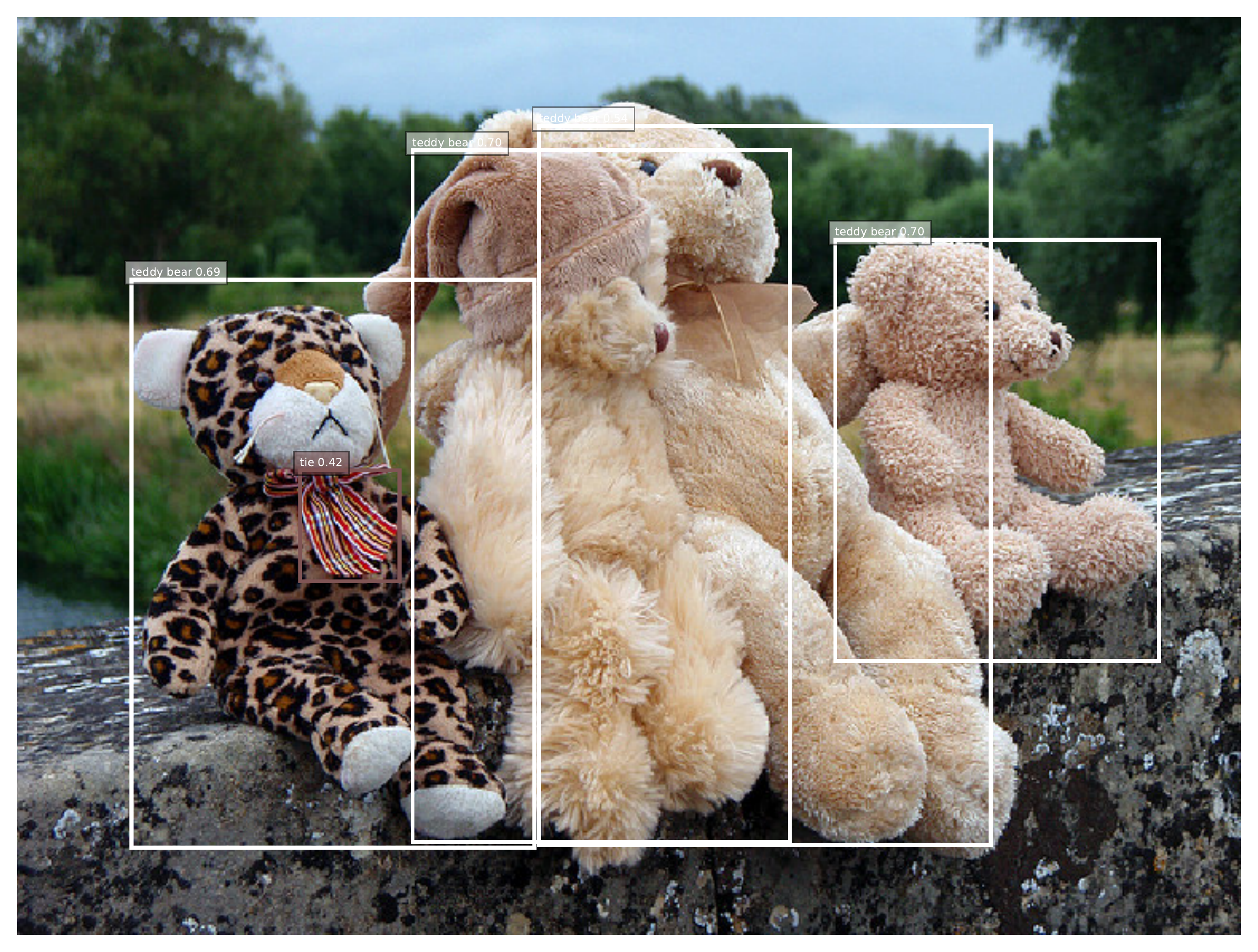}
\end{subfigure}
\vspace{-7mm}
  \caption{Some detection examples. Top: Baseline results by RetinaNet with focal loss. Bottom: Our results with AP-loss.}
\vspace{-2mm}
\label{fig-result}
\end{figure*}

\begin{table*}[t]
\footnotesize
\centering
\setlength{\tabcolsep}{3mm}{
\resizebox{1\linewidth}{!}{%
\begin{tabular}{c|c|c|c|c|ccc|ccc}
\hline
\multirow{2}*{Method} & \multirow{2}*{Backbone} & \multirow{2}*{Multi-Scale} & VOC07 & VOC12 & \multicolumn{6}{c}{COCO} \\
\cline{4-11}
 & & & AP\(_{50}\) & AP\(_{50}\) & AP & AP\(_{50}\) & AP\(_{75}\) & AP\(_S\) & AP\(_M\) & AP\(_L\) \\
\hline\hline
YOLOv2~\cite{redmon2017yolo9000} & DarkNet-19 & \xmark & 78.6 & 73.4 & 21.6 & 44.0 & 19.2 & 5.0 & 22.4 & 35.5 \\
DSOD300~\cite{shen2017dsod} & DS/64-192-48-1 & \xmark & 77.7 & 76.3 & 29.3 & 47.3  & 30.6  & 9.4  & 31.5  & 47.0  \\
SSD512~\cite{liu2016ssd} & VGG-16 & \xmark & 79.8 & 78.5 & 28.8 & 48.5 & 30.3 & - & - & - \\
SSD513~\cite{fu2017dssd} & ResNet-101 & \xmark & 80.6 & 79.4 & 31.2 & 50.4 & 33.3 & 10.2 & 34.5 & 49.8 \\
DSSD513~\cite{fu2017dssd} & ResNet-101 & \xmark & 81.5 & 80.0 & 33.2 & 53.3 & 35.2 & 13.0 & 35.4 & 51.1 \\
DES512~\cite{Zhang_2018_CVPR} & VGG-16 & \xmark & 81.7 & 80.3 & 32.8 & 53.2 & 34.6 & 13.9 & 36.0 & 47.6 \\
RFBNet512~\cite{Liu_2018_ECCV} & VGG-16 & \xmark & 82.2 & - & 33.8 & 54.2 & 35.9 & 16.2 & 37.1 & 47.4 \\
PFPNet-R512~\cite{kim2018parallel} & VGG-16 & \xmark & 82.3 & 80.3 & 35.2 & 57.6 & 37.9 & \textbf{18.7} & 38.6 & 45.9 \\
RefineDet512~\cite{zhang2018single} & VGG-16 & \xmark & 81.8 & 80.1 & 33.0 & 54.5 & 35.5 & 16.3 & 36.3 & 44.3 \\
RefineDet512~\cite{zhang2018single} & ResNet-101 & \xmark & - & - & 36.4 & 57.5 & 39.5 & 16.6 & 39.9 & 51.4 \\
RetinaNet500~\cite{lin2018focal} & ResNet-101 & \xmark & - & - & 34.4 & 53.1 & 36.8 & 14.7 & 38.5 & 49.1 \\
RetinaNet500+AP-Loss (ours) & ResNet-101 & \xmark & \textbf{83.9} & \textbf{83.1} & \textbf{37.4} & \textbf{58.6} & \textbf{40.5} & 17.3 & \textbf{40.8} & \textbf{51.9} \\
\cline{1-11}
PFPNet-R512~\cite{kim2018parallel} & VGG-16 & \cmark & 84.1 & 83.7 & 39.4 & 61.5 & 42.6 & 25.3 & 42.3 & 48.8 \\
RefineDet512~\cite{zhang2018single} & VGG-16 & \cmark & 83.8 & 83.5 & 37.6 & 58.7 & 40.8 & 22.7 & 40.3 & 48.3 \\
RefineDet512~\cite{zhang2018single} & ResNet-101 & \cmark & - & - & 41.8 & 62.9 & 45.7 & \textbf{25.6} & \textbf{45.1} & \textbf{54.1} \\
RetinaNet500+AP-Loss (ours) & ResNet-101 & \cmark & \textbf{84.9} & \textbf{84.5} & \textbf{42.1} & \textbf{63.5} & \textbf{46.4} & \textbf{25.6} & 45.0 & 53.9 \\
\hline
\end{tabular}}
}
\vspace{-3mm}
\caption{Detection results on VOC2007 {\tt test}, VOC 2012 {\tt test} and COCO {\tt test-dev} sets.}
\label{state-of-the-art}
\vspace{-4mm}
\end{table*}

\subsubsection{Comparison on Different Optimization Methods}
We also compare our optimization method with the approximate gradient method~\cite{song2016training,henderson2016end} and structured hinge loss method~\cite{Mohapatra_2018_CVPR}. Both~\cite{song2016training,henderson2016end} approximate the AP-loss with a smooth expectation and envelope function, respectively. Following their guidance, we replace the step function in AP-loss with a sigmoid function to constrain the gradient to neither zero nor undefined, while still keep the shape similar to the original function. Same as~\cite{henderson2016end}, we adopt the log space objective function, \textit{i.e.} \(log(\text{AP}+\epsilon)\), to allow the model to quickly escape from the initial state. We train the detector on VOC2007 {\tt trainval} set and turn off the bounding box regression task. The convergence curves shown in \autoref{fig-convergence_curve} reveal some essential observations. It can be seen that AP-loss optimized by approximate gradient method does not even converge, likely because its non-convexity and non-quasiconvexity fail on a direct gradient descent method. Meanwhile, AP-loss optimized by the structured hinge loss method~\cite{Mohapatra_2018_CVPR} converges slowly and stabilizes near 0.8, which is significantly worse than the asymptotic limit of AP-loss optimized by our error-driven update scheme. We believe that this method does not optimize the AP-loss directly but rather an upper bound of it, which is controlled by a discriminant function~\cite{Mohapatra_2018_CVPR}. In ranking task, this discriminant function is hand-picked and has an AUC-like form, which may cause variability in optimization.

\subsection{Benchmark Results}
With the settings selected in ablation study, we conduct experiments to compare the proposed detector to state-of-the-art one-stage detectors on three widely used benchmark, \textit{i.e.} VOC2007 {\tt test}, VOC2012 {\tt test} and COCO {\tt test-dev} sets. We use ResNet-101 as backbone networks instead of ResNet-50 in ablation study. We use an image scale of 500 pixels for testing. \autoref{state-of-the-art} lists the benchmark results comparing to recent state-of-the-art one-stage detectors such as SSD~\cite{liu2016ssd}, YOLOv2~\cite{redmon2017yolo9000}, DSSD~\cite{fu2017dssd}, DSOD~\cite{shen2017dsod}, DES~\cite{Zhang_2018_CVPR}, RetinaNet~\cite{lin2018focal}, RefineDet~\cite{zhang2018single}, PFPNet~\cite{kim2018parallel}, RFBNet~\cite{Liu_2018_ECCV}. Compared to the baseline model RetinaNet500~\cite{lin2018focal}, our detector achieves a 3.0\% improvement (37.4\% \textit{vs.} 34.4\%) on COCO dataset. \autoref{fig-result} illustrates some detection results by the RetinaNet with focal loss and our AP-loss. Besides, our detector outperforms all the other methods for both single-scale and multi-scale tests in all the three benchmarks. We should emphasize that this verifies the great effectiveness of our AP-loss since our detector achieves such a great performance gain just by replacing the focal-loss with our AP-loss in RetinaNet without whistle and bells, without using advanced techniques like deformable convolution~\cite{dai2017deformable}, SNIP~\cite{singh2018snip}, group normalization~\cite{wu2018group}, etc. The performance could be further improved with these kinds of techniques and other possible tricks. Our detector has the same detection speed (\textit{i.e.}, $\sim$$11~fps$ on one NVidia TitanX GPU) as RetinaNet500~\cite{lin2018focal} since it does not change the network architecture for inference.

\section{Conclusion}
In this paper, we address the class imbalance issue in one-stage object detectors by replacing the classification sub-task with a ranking sub-task, and proposing to solve the ranking task with AP-Loss. Due to non-differentiability and non-convexity of the AP-loss, we propose a novel algorithm to optimize it based on error-driven update scheme from perceptron learning. We provide a grounded theoretical analysis of the proposed optimization algorithm. Experimental results show that our approach can significantly improve the state-of-the-art one-stage detectors.

{\bf\noindent Acknowledgements.}
This paper is supported in part by: National Natural Science Foundation of China (61471235), Shanghai 'The Belt and Road' Young Scholar Exchange Grant (17510740100), CREST Malaysia (No. T03C1-17), and the PKU-NTU Joint Research Institute (JRI) sponsored by a donation from the Ng Teng Fong Charitable Foundation. We gratefully acknowledge the support from Tencent YouTu Lab.

{\small
\bibliographystyle{ieee_fullname}
\bibliography{egbib}
}

\section*{A1. Convergence}
We provide proof for the proposition mentioned in Section 3.3.1 of the paper. The proof is generalized from the original convergence proof~\cite{Novikoff1963ON} for perceptron learning algorithm.
\begin{proposition}
The AP-loss optimizing algorithm is guaranteed to converge in finite steps if below conditions hold:
{\noindent (1) the learning model is linear;} \par
{\noindent (2) the training data is linearly separable.}
\end{proposition}

{\noindent\textit{Proof.}}
Let \(\bm{\theta}\) denote the weights of the linear model. Let \(\bm{f}_{k}^{(n)}\) denote the feature vector of \(k\)-th box in \(n\)-th training sample. Assume the number of training samples is finite and each training sample contains at most \(M\) boxes. Hence the score of \(k\)-th box is \(s_{k}^{(n)}=\langle \bm{f}_{k}^{(n)} , \bm{\theta} \rangle\). Define \(x_{ij}^{(n)}=-(s_{i}^{(n)}-s_{j}^{(n)})\). Note that the training data is separable, which means there are \(\epsilon>0\) and \(\bm{\theta}^{*}\) that satisfy:
\begin{equation}
\small
\forall n, \,\, \forall i\in \mathcal{P}^{(n)}, \,\, \forall j\in \mathcal{N}^{(n)}, \,\, \langle \bm{f}_{i}^{(n)} , \bm{\theta}^{*} \rangle \geq \langle \bm{f}_{j}^{(n)}, \bm{\theta}^{*}\rangle + \epsilon
\end{equation}

In the \(t\)-th step, a training sample which makes an error (if there is no such training sample, the model is already optimal and algorithm will stop) is randomly chosen. Then the update of \(\bm{\theta}\) is:
\begin{equation}
\small
\bm{\theta}^{(t+1)}=\bm{\theta}^{(t)}+\sum_{i\in \mathcal{P}}\sum_{j\in \mathcal{N}}L_{ij}(\bm{x})\cdot (\bm{f}_i-\bm{f}_j)
\label{update}
\end{equation}
where
\begin{equation}
\small
L_{ij}(\bm{x})=\frac{H(x_{ij})}{1+\sum_{k\neq i} H(x_{ik})}
\end{equation}
Here, since the discussion centers on the current training sample, we omit the superscript hereon.

From (\ref{update}), we have
\begin{equation}
\small
\begin{split}
\langle \bm{\theta}^{(t+1)}, \bm{\theta}^{*} \rangle &=\langle \bm{\theta}^{(t)},\bm{\theta}^{*}\rangle+\sum_{i\in \mathcal{P}}\sum_{j\in \mathcal{N}}L_{ij}\langle (\bm{f}_i-\bm{f}_j),\bm{\theta}^{*} \rangle \\
&\geq \langle \bm{\theta}^{(t)},\bm{\theta}^{*}\rangle+\sum_{i\in \mathcal{P}}\sum_{j\in \mathcal{N}}L_{ij}\epsilon \\
&\geq \langle \bm{\theta}^{(t)},\bm{\theta}^{*}\rangle+\max_{i\in \mathcal{P},j\in \mathcal{N}}\{L_{ij}\}\epsilon \\
&\geq \langle \bm{\theta}^{(t)},\bm{\theta}^{*} \rangle + \frac{1}{|\mathcal{P}|+|\mathcal{N}|}\epsilon \\
&\geq \langle \bm{\theta}^{(t)},\bm{\theta}^{*} \rangle + \frac{1}{M}\epsilon
\end{split}
\end{equation}
For convenience, let \(\bm{\theta}^{(0)}=0\) (if \(\bm{\theta}^{(0)}\neq0\), we can still find a \(c>0\) that satisfies (\ref{lowerbound}) for sufficiently large \(t\)), we have
\begin{equation}
\small
\langle \bm{\theta}^{(t)},\bm{\theta}^{*} \rangle \geq \frac{1}{M}\epsilon \cdot t
\end{equation}
Then
\begin{equation}
\small
\|\bm{\theta}^{(t)}\| \geq \frac{\langle \bm{\theta}^{(t)},\bm{\theta}^{*} \rangle}{\|\bm{\theta}^{*}\|} \geq \frac{1}{M\cdot \|\bm{\theta}^{*}\|}\epsilon \cdot t \geq c \cdot t
\label{lowerbound}
\end{equation}
Here, \(c\) is a positive constant.

From (\ref{update}), we also have
\begin{equation}
\small
\begin{split}
&\|\bm{\theta}^{(t+1)}\|^2 \\
= & \|\bm{\theta}^{(t)}\|^2+\|\sum_{i\in \mathcal{P}}\sum_{j\in \mathcal{N}}L_{ij}(\bm{f}_i-\bm{f}_j)\|^2 \\
& \,\,\,\,\,\,\,\,\,\,\,\,\,\,\,\,\,\,\,\,\,\,\,\,\,\,\,\,\,\,\,\,\,\,\,\,\,\,\,\,\,\,\,\,\,\,\, +2\langle \sum_{i\in \mathcal{P}}\sum_{j\in \mathcal{N}}L_{ij}(\bm{f}_i-\bm{f}_j),\bm{\theta}^{(t)} \rangle \\
= & \|\bm{\theta}^{(t)}\|^2 + \|\sum_{i\in \mathcal{P}}\sum_{j\in \mathcal{N}}L_{ij}(\bm{f}_i-\bm{f}_j)\|^2+2\sum_{i\in \mathcal{P}}\sum_{j\in \mathcal{N}}L_{ij} x_{ji} \\
\leq & \|\bm{\theta}^{(t)}\|^2 + \|\sum_{i \in \mathcal{P}}\sum_{j\in \mathcal{N}} L_{ij}(\bm{f}_i-\bm{f}_j)\|^2 \\
\leq & \|\bm{\theta}^{(t)}\|^2 + |\mathcal{P}|\cdot |\mathcal{N}| \cdot \max_{i\in \mathcal{P},j\in \mathcal{N}}\{\|\bm{f}_i-\bm{f}_j\|^2\} \\
\leq & \|\bm{\theta}^{(t)}\|^2 + M^2\cdot \max_{n,i,j}\{\|\bm{f}^{(n)}_i-\bm{f}^{(n)}_j\|^2\} \\
\leq & \|\bm{\theta}^{(t)}\|^2 + C
\end{split}
\end{equation}
Here, \(C\) is a positive constant. Let \(\bm{\theta}^{(0)}=0\) (again, if \(\bm{\theta}^{(0)}\neq0\), we can still find a \(C>0\) that satisfies (\ref{upperbound}) for sufficiently large \(t\)), we arrive at:
\begin{equation}
\small
\|\bm{\theta}^{(t)}\|^2 \leq C \cdot t
\label{upperbound}
\end{equation}
Then, combining (\ref{lowerbound}) and (\ref{upperbound}), we have
\begin{equation}
\small
c^2\cdot t^2\leq \|\bm{\theta}^{(t)}\|^2 \leq C\cdot t
\end{equation}
which means
\begin{equation}
\small
t \leq \frac{C}{c^2}
\end{equation}
It shows that the algorithm will stop at most after \(C/c^2\) steps, which means that the training model will achieve the optimal solution in finite steps.

\section*{A2. An Example of Gradient Descent Failing on Smoothed AP-loss}

We approximate the step function in AP-loss by sigmoid function to make it amenable to gradient descent. Specifically, the smoothed AP-loss function is given by:
\begin{equation}
\small
F =\frac{1}{|\mathcal{P}|}\sum_{i\in \mathcal{P}}\sum_{j\in \mathcal{N}}\frac{S(x_{ij})}{1+\sum_{k\neq i}S(x_{ik})}
\end{equation}
where
\begin{equation}
\small
S(x)=\frac{e^x}{1+e^x}
\end{equation}
Consider a linear model \(s=f_1 \theta_1+f_2 \theta_2\) and three training samples \((0,0),(1,0),(-3,1)\) (the first one is negative sample, others are positive samples). Then we have
\begin{equation}
\small
\begin{split}
&s^{(1)}=0\cdot \theta_1+0 \cdot \theta_2 \\
&s^{(2)}=1\cdot \theta_1+0\cdot \theta_2 \\
&s^{(3)}=-3\cdot \theta_1 + 1 \cdot \theta_2
\end{split}
\end{equation}
Note that the training data is separable since we have \(s^{(2)} > s^{(1)}\) and \(s^{(3)} > s^{(1)}\) when \(0< \theta_1 < \frac{1}{3} \cdot \theta_2 \).

Under this setting, the smoothed AP-loss become
\begin{equation}
\small
\begin{split}
& F(\theta_1,\theta_2)=\frac{1}{2}(\frac{S(-\theta_1)}{1+S(-\theta_1)+S(\theta_2-4\theta_1)} \\
& \,\,\,\,\,\,\,\,\,\,\,\,\,+\frac{S(3\theta_1-\theta_2)}{1+S(4\theta_1-\theta_2)+S(3\theta_1-\theta_2)})
\end{split}
\end{equation}
If \(\theta_1\) is sufficiently large and \(\theta_1>\theta_2>0\), then the partial derivatives satisfy the following condition:
\begin{equation}
\small
\frac{\partial F}{\partial \theta_1}<\frac{\partial F}{\partial \theta_2}<0
\label{partial}
\end{equation}
which means \(\theta_1\) and \(\theta_2\) will keep increasing with the inequality \(\theta_1>\theta_2\) according to the gradient descent algorithm. Hence the objective function \(F\) will approach \(1/6\) here. However, the objective function \(F\) approaches the global minimum value \(0\) if and only if \(\theta_1 \rightarrow +\infty\) and \( \theta_2 - 3 \theta_1 \rightarrow +\infty\). This shows that the gradient descent fails to converge to global minimum in this case.

\section*{A3. Inseparable Case}

In this section, we will provide analysis for our algorithm with inseparable training data. We demonstrate that the bound of accumulated AP-loss depends on the best performance of learning model. The analysis is based on online learning bounds~\cite{shalev2012online}.

\subsection*{A3.1. Preliminary}
To handle the inseparable case, a mild modification on the proposed algorithm is needed, \textit{i.e.} in the error-driven update scheme, \(L_{ij}\) is modified to
\begin{equation}
\small
\widetilde{L}_{ij}=\frac{\widetilde{H}(x_{ij})}{1+\sum_{k\in \mathcal{P}\cup\mathcal{N},k\neq i}H(x_{ik})}
\end{equation}
where \(\widetilde{H}(\cdot)\) is defined in Section 3.4.2 (Piecewise Step Function) of the paper. The purpose is to introduce a non-zero decision margin for the pairwise score \(x_{ij}\) which makes the algorithm more robust in the inseparable case. In contrast to the case in Section 3.4.2, here we only change \(H(\cdot)\) to \(\widetilde{H}(\cdot)\) in the numerator for the convenience of theoretical anaysis. However, such algorithm still suffers from the discontinuity of \(H(\cdot)\) in the denominator. Hence the strategy in Section 3.4.2 is also practical consideration, necessary for good performance. Then, consider the AP-loss:
\begin{equation}
\small
\mathcal{L}_{AP}(\bm{x};\mathcal{P},\mathcal{N})=\frac{1}{|\mathcal{P}|}\sum_{i\in \mathcal{P}}\frac{\sum_{j\in \mathcal{N}} H(x_{ij})}{1+\sum_{j\in \mathcal{P} \cup\mathcal{N},j\neq i}H(x_{ij})}
\end{equation}
and define a surrogate loss function:
\begin{equation}
\small
l(\bm{x},\hat{\bm{x}};\mathcal{P},\mathcal{N})=\frac{1}{|\mathcal{P}|}\sum_{i\in \mathcal{P}}\frac{\sum_{j\in \mathcal{N}}Q(x_{ij})}{1+\sum_{j\in \mathcal{P}\cup\mathcal{N},j \neq i}H(\hat{x}_{ij})}
\end{equation}
where \(Q(x)=\int_{-\infty}^{x}{\widetilde{H}(\upsilon)d\upsilon}\). Note that the AP-loss is upper bounded by the surrogate loss:
\begin{equation}
\small
l(\bm{x},\bm{x};\mathcal{P},\mathcal{N})\geq \frac{\delta}{4} \mathcal{L}_{AP}(\bm{x};\mathcal{P},\mathcal{N})\
\end{equation}
The learning model can be written as \(\bm{x}=\bm{X}_{d}(\bm{\theta})\), where \(d\in \mathcal{D}\) denotes the training data for one iteration and \(D\) is the whole training set. Then, the modified error-driven algorithm is equivalent to gradient descent on surrogate loss \(l(\bm{X}_{d^{(t)}}(\bm{\theta}),\bm{X}_{d^{(t)}}(\bm{\theta}^{(t)});\mathcal{P}_{d^{(t)}},\mathcal{N}_{d^{(t)}})\) at each step \(t\).
We further suppose below conditions are satisfied:\par
\noindent(1) For all \(\hat{\bm{\theta}}\) and \(d \in \mathcal{D}\), \(l(\bm{X}_{d}(\bm{\theta}),\bm{X}_{d}(\hat{\bm{\theta}});\mathcal{P}_d,\mathcal{N}_d)\) is convex \textit{w.r.t} \(\bm{\theta}\).\par
\noindent(2) For all \(d \in \mathcal{D}\), \(\|\partial \bm{X}_{d}(\bm{\theta}) / \partial \bm{\theta}\|\) is upper bounded by a constant \(R\). Here \(\|\cdot\|\) is the matrix norm induced by the 2-norm for vectors.

\noindent \textit{\textbf{Remark 1.}} Note that these two conditions are satisfied if the learning model is linear.\par

\subsection*{A3.2. Bound of Accumulated Loss}

By the convexity, we have:
\begin{equation}
\small
l^{(t)}(\bm{\theta})\leq l^{(t)}(\bm{u})+\langle\bm{\theta}-\bm{u},\frac{\partial l^{(t)}(\bm{\theta})}{\partial \bm{\theta}}\rangle
\end{equation}
where \(l^{(t)}(\bm{\theta})\) denotes \(l(\bm{X}_{d^{(t)}}(\bm{\theta}),\bm{X}_{d^{(t)}}(\bm{\theta}^{(t)});\mathcal{P}_{d^{(t)}},\mathcal{N}_{d^{(t)}})\) and \(\bm{u}\) can be any vector of model weights.
Then, let \(\bm{\theta}=\bm{\theta}^{(t)}\) and compute the sum over \(t=1 \sim T\), we have:
\begin{equation}
\small
\begin{split}
&\sum_{t=1}^{T} l^{(t)}(\bm{\theta}^{(t)}) - \sum_{t=1}^{T}l^{(t)}(\bm{u})
\leq \sum_{t=1}^{T}\langle \bm{\theta}^{(t)}-\bm{u},\frac{\partial l^{(t)}(\bm{\theta}^{(t)})}{\partial \bm{\theta}}\rangle \\
&=\sum_{t=1}^{T} \langle \bm{\theta}^{(t)}-\bm{u},\frac{1}{\eta}(\bm{\theta}^{(t)}-\bm{\theta}^{(t+1)})\rangle \\
&\leq \frac{1}{2\eta} \|\bm{u}-\bm{\theta}^{(1)}\|^2 +
\frac{1}{2\eta} \sum_{t=1}^{T} \|\bm{\theta}^{(t)}-\bm{\theta}^{(t+1)}\|^2\\
&=\frac{1}{2\eta} \|\bm{u}-\bm{\theta}^{(1)}\|^2 + \frac{\eta}{2}\sum_{t=1}^{T}\|\frac{\partial l^{(t)}(\bm{\theta}^{(t)})}{\partial \bm{\theta}}\|^2
\end{split}
\end{equation}
where \(\eta\) is the step size of gradient descent. Note that
\begin{equation}
\small
\frac{\partial l^{(t)}(\bm{\theta}^{(t)})}{\partial \bm{\theta}}=\frac{\partial \bm{X}(\bm{\theta})}{\partial {\bm{\theta}}} \big|_{\bm{\theta}=\bm{\theta}^{(t)}} \cdot
\frac{\partial l(\bm{x},\bm{x}^{(t)})}{\partial \bm{x}} \big|_{\bm{x}=\bm{X}(\bm{\theta}^{(t)})}
\end{equation}
and
\begin{equation}
\small
\begin{split}
&\|\frac{\partial l(\bm{x},\bm{x}^{(t)})}{\partial \bm{x}}\|^{2}
=\frac{1}{|\mathcal{P}|^{2}}\sum_{i\in \mathcal{P}}\frac{\sum_{j\in \mathcal{N}} \widetilde{H}^{2}(x_{ij})}{(1+\sum_{j\in \mathcal{P}\cup\mathcal{N},j\neq i} H(x^{(t)}_{ij}))^{2}}\\
&\leq \frac{1}{|\mathcal{P}|^{2}}\sum_{i\in \mathcal{P}} \frac{\frac{1}{\delta} \sum_{j\in \mathcal{N}} Q(x_{ij})}{1+\sum_{j\in \mathcal{P}\cup\mathcal{N},j\neq i} H(x^{(t)}_{ij})}\leq\frac{1}{\delta } l(\bm{x},\bm{x}^{(t)})\\
\end{split}
\end{equation}
Note that both \(\mathcal{P}_{d}\) and \(\mathcal{N}_{d}\) depend on \(d\). However, we omit the subscript \(d\) here since the discussion only centers on the current training sample \(d^{(t)}\).

Hence we have:
\begin{equation}
\small
\begin{split}
&\sum_{t=1}^{T} l^{(t)}(\bm{\theta}^{(t)}) - \sum_{t=1}^{T}l^{(t)}(\bm{u})\\
&\leq \frac{1}{2\eta} \|\bm{u}-\bm{\theta}^{(1)}\|^2 + \frac{\eta R^2}{2 \delta} \sum_{t=1}^{T} l^{(t)}(\bm{\theta}^{(t)}).
\end{split}
\end{equation}
Let \(\eta=\delta/R^2\), rearrange and get the expression:
\begin{equation}
\small
\frac{1}{2}\sum_{t=1}^{T}l^{(t)}(\bm{\theta}^{(t)})\leq \frac{R^2}{2\delta} \|\bm{u}-\bm{\theta}^{(1)}\|^2 + \sum_{t=1}^{T}l^{(t)}(\bm{u})
\end{equation}
This entails the bound of surrogate loss \(l\):
\begin{equation}
\small
\begin{split}
\sum_{t=1}^{T}l^{(t)}(\bm{\theta}^{(t)})\leq 2\sum_{t=1}^{T}l^{(t)}(\bm{u})+ \frac{R^2}{\delta} \|\bm{u}-\bm{\theta}^{(1)}\|^2
\end{split}
\end{equation}
which implies the bound of AP-loss \(\mathcal{L}_{AP}\):
\begin{equation}
\small
\sum_{t=1}^{T} \mathcal{L}_{AP}(\bm{X}(\bm{\theta}^{(t)})) \leq \frac{8}{\delta}\sum_{t=1}^{T}l^{(t)}(\bm{u})+ \frac{4R^2}{\delta^2} \|\bm{u}-\bm{\theta}^{(1)}\|^2
\label{regret_bound1}
\end{equation}
As a special case, if there exists a \(\bm{u}\) such that \(l^{(t)}(\bm{u})=0\) for all \(t\), then the accumulated AP-loss is bounded by a constant, which implies that convergence can be achieved with finite steps (similar to that of the separable case). Otherwise, with sufficiently large \(T\), the average AP-loss mainly depends on \(\frac{1}{T}\frac{8}{\delta}\sum_{t=1}^{T}l^{(t)}(\bm{u})\). This implies that the bound is meaningful if there still exists a sufficiently good solution \(\bm{u}\) in such inseparable case.

\subsection*{A3.3. Offline Setting}
With the offline setting (\(d^{(t)}=d\) for all \(t\)), a bound with simpler form can be revealed. For simplicity, we will omit the subscript \(d\) of \(\bm{X}_{d}(\bm{u}),\mathcal{P}_d, \mathcal{N}_d\) and define \(A_{i}(\bm{u})=\sum_{j\in\mathcal{N}}Q(X_{ij}(\bm{u}))\), \(Z(\bm{u})=\max_{i\in \mathcal{P}}\{A_{i}(\bm{u})\}\).
Then,
\begin{equation}
\small
\begin{split}
&l^{(t)}(\bm{u})= \frac{1}{|\mathcal{P}|}\sum_{i\in \mathcal{P}}\frac{\sum_{j\in \mathcal{N}}Q(X_{ij}(\bm{u}))}{1+\sum_{j\in \mathcal{P}\cup\mathcal{N},j\neq i}H(X_{ij}(\bm{\theta}^{(t)}))}\\
&= \frac{1}{|\mathcal{P}|}\sum_{i\in \mathcal{P}}\frac{A_i(\bm{u})}{1+\sum_{j\in \mathcal{P}\cup\mathcal{N},j\neq i}H(X_{ij}(\bm{\theta}^{(t)}))}\\
&\leq \frac{Z(\bm{u})}{|\mathcal{P}|}\sum_{i=1}^{|\mathcal{P}|}\frac{1}{i}\leq \frac{\ln|\mathcal{P}|+1}{|\mathcal{P}|}Z(\bm{u})
\end{split}
\label{bound1}
\end{equation}
The second last inequality is based on the fact that \((1+\sum_{j\in \mathcal{P}\cup\mathcal{N},j\neq i}H(X_{ij}(\bm{\theta}^{(t)})))\) are picked from \(1\sim (|\mathcal{P}|+|\mathcal{N}|)\) without replacement (assume no ties; if ties exist, this inequality still holds).
Combining the results from \autoref{bound1} and \autoref{regret_bound1}, we have:
\begin{equation}
\footnotesize
\frac{1}{T}\sum_{t=1}^{T}\mathcal{L}_{AP}(\bm{X}(\bm{\theta}^{(t)}))\leq \frac{\ln|\mathcal{P}|+1}{|\mathcal{P}|}\cdot\frac{8}{\delta}Z(\bm{u})+ \frac{1}{T}\frac{4R^2\|\bm{u}-\bm{\theta}^{(1)}\|^2}{\delta^2}
\label{final_bound1}
\end{equation}
Next,
\begin{equation}
\footnotesize
\begin{split}
&l^{(t)}(\bm{u})=\frac{1}{|\mathcal{P}|}\sum_{i\in \mathcal{P}}\frac{A_i(\bm{u})}{1+\sum_{j\in \mathcal{P}\cup\mathcal{N},j\neq i}H(X_{ij}(\bm{\theta}^{(t)}))} \\
&=\frac{1}{|\mathcal{P}|}\sum_{i\in \mathcal{P}}\frac{1+\sum_{j\in\mathcal{P},j\neq i}H(X_{ij}(\bm{\theta}^{(t)}))}{1+\sum_{j\in \mathcal{P}\cup\mathcal{N},j\neq i}H(X_{ij}(\bm{\theta}^{(t)}))}\\
&\quad\quad\quad\quad\quad\quad\quad\quad\quad\quad\quad \cdot\frac{A_i(\bm{u})}{1+\sum_{j\in\mathcal{P},j\neq i}H(X_{ij}(\bm{\theta}^{(t)}))}\\
&\leq \frac{1}{|\mathcal{P}|}\sum_{i\in \mathcal{P}}\frac{1+\sum_{j\in\mathcal{P},j\neq i}H(X_{ij}(\bm{\theta}^{(t)}))}{1+\sum_{j\in \mathcal{P}\cup\mathcal{N},j\neq i}H(X_{ij}(\bm{\theta}^{(t)}))}\cdot Z(\bm{u})\\
&=(1-\mathcal{L}_{AP}(\bm{X}(\bm{\theta}^{(t)})))\cdot Z(\bm{u})
\end{split}
\label{bound2}
\end{equation}
Combining the results from \autoref{bound2} and \autoref{regret_bound1}, we have:
\begin{equation}
\footnotesize
\frac{1}{T}\sum_{t=1}^{T}\mathcal{L}_{AP}(\bm{X}(\bm{\theta}^{(t)}))\leq \frac{\frac{8}{\delta}Z(\bm{u})}{1+\frac{8}{\delta}Z(\bm{u})}+ \frac{1}{T}\frac{4R^2\|\bm{u}-\bm{\theta}^{(1)}\|^2}{\delta^2}
\label{final_bound2}
\end{equation}
If \(Z(\bm{u})\) is small, the bound in \autoref{final_bound1} is active, otherwise the bound in \autoref{final_bound2} is active. Consequently, we have:
\begin{equation}
\footnotesize
\overline{\mathcal{L}_{AP}}\leq \min\{\frac{\ln|\mathcal{P}|+1}{|\mathcal{P}|}\frac{8}{\delta}Z(\bm{u}),\frac{\frac{8}{\delta}Z(\bm{u})}{1+\frac{8}{\delta}Z(\bm{u})}\}+\epsilon
\end{equation}
where \(\overline{\mathcal{L}_{AP}}\) denotes the average AP-loss, \(\epsilon\rightarrow 0\) as \(T\) increases.

\section*{A4. Consistency}
\begin{observation}
When the activation function \(L(\cdot)\) takes the form of softmax function and loss-augmented step function, our optimization algorithm can be expressed as the gradient descent algorithm on cross-entropy loss and hinge loss respectively.
\label{consistency}
\end{observation}

{\bf \noindent Cross Entropy Loss:}
Consider the multi-class classification task. The outputs of neural network are \((x_1,\ldots, x_K)\) where \(K\) is the number of classes, and the ground truth label is \(y \in \{1,\ldots,K\}\). Using softmax as the activation function, we have:
\begin{equation}
\small
\begin{split}
&(L_1,\ldots,L_K) \\
&= softmax(\bm{x}) = (\frac{e^{x_1}}{\sum_i e^{x_i}},\ldots,\frac{e^{x_K}}{\sum_i e^{x_i}})
\end{split}
\end{equation}
The cross entropy loss is:
\begin{equation}
\small
\mathcal{L}_{ce}=-\sum_i \mathbf{1}_{y=i} \log(L_i)
\end{equation}
Hence the gradient of \(x_i\) is
\begin{equation}
\small
g_i=L_i-\mathbf{1}_{y=i}
\end{equation}
Note that \(g_i\) is ``error-driven'' with the desired output \(\mathbf{1}_{y=i}\) and current output \(L_i\). This form is consistent with our error-driven update scheme (c.f. Section 3.2.1 of the paper).

{\bf \noindent Hinge Loss:}
Consider the binary classification task. The output of neural network is \(x\), and the ground truth label is \(y \in \{1,2\}\). Define \((x_1,x_2)=(-x,x)\). Using loss-augmented step function as the activation function, we have:
\begin{equation}
\small
(L_1,L_2)=(H(x_1-1),H(x_2-1))
\end{equation}
where \(H(\cdot)\) is the Heaviside step function. The hinge loss is:
\begin{equation}
\small
\mathcal{L}_{hinge}=\mathbf{1}_{y=1} \max\{1-x_1,0\} + \mathbf{1}_{y=2} \max\{1-x_2,0\}
\end{equation}
Hence the gradient of \(x_i\) is
\begin{equation}
\small
g_i = \mathbf{1}_{y=i} \cdot (L_i-1)
\end{equation}
There are two cases. If \(y=i\), the gradient \(g_i\) is ``error-driven'' with the desired output \(1\) and current output \(L_i\). If \(y \neq i\), the gradient \(g_i\) equals zero, since \(x_i\) does not contribute to the loss. This form is consistent with our error-driven update scheme (c.f. Section 3.2.1 of the paper).

\section*{A5. Additional Experiments on SSD}
\subsection*{A5.1. Experimental Settings}
We also evaluate the proposed AP-loss on another one-stage detector SSD~\cite{liu2016ssd}. The models are trained on VOC2007 and VOC2012 {\tt trainval} sets, and tested on VOC2007 {\tt test} set. We use VGG-16~\cite{simonyan2014very} as the backbone model which is pre-trained on the ImageNet-1k classification dataset~\cite{deng2009imagenet}. We use {\tt conv4\_3}, {\tt conv7}, {\tt conv8\_2}, {\tt conv9\_2}, {\tt conv10\_2}, {\tt conv11\_2}, {\tt conv12\_2} to predict both location and their corresponding confidences. An additional convolution layer is added after {\tt conv4\_3} to scale the feature. The associated anchors are the same as that designed in~\cite{liu2016ssd}. In testing phase, the input image is fixed to 512\(\times\)512. For focal loss with SSD, we observe that the hyper-parameters \(\gamma=1,\alpha=0.25\) lead to a much better performance than the original settings in~\cite{lin2018focal} which are \(\gamma=2,\alpha=0.25\). Hence we evaluate the focal loss with new \(\gamma\) and \(\alpha\) in our experiments on SSD. Other details are similar to that in Section 4.1 of the paper.

\subsection*{A5.2. Results}

\begin{table}[t]
\small
\centering
\setlength{\tabcolsep}{4mm}{
\begin{tabular}{c|ccc}
\hline
\multirow{2}*{Training Loss} & \multicolumn{3}{c}{PASCAL VOC} \\
\cline{2-4}
 & AP & AP\(_{50}\) & AP\(_{75}\) \\
\hline\hline
CE-Loss + OHEM & 43.6 & 76.0 & 44.7\\
Focal Loss & 39.3 & 69.9 & 38.0 \\
\cline{1-4}
AUC-Loss & 33.8 & 63.7 & 31.5 \\
AP-Loss & \textbf{45.2} & \textbf{77.3} & \textbf{47.3} \\
\hline
\end{tabular}}
\vspace{-2mm}
\caption{Comparison through different training losses. Models are tested on VOC2007 {\tt test} set. The metric AP is averaged over multiple IoU thresholds of \(0.50:0.05:0.95\).}
\vspace{-4mm}
\label{base-loss-ssd}
\end{table}

\begin{figure}[t]
\centering
\small
\includegraphics[width=0.8\linewidth]{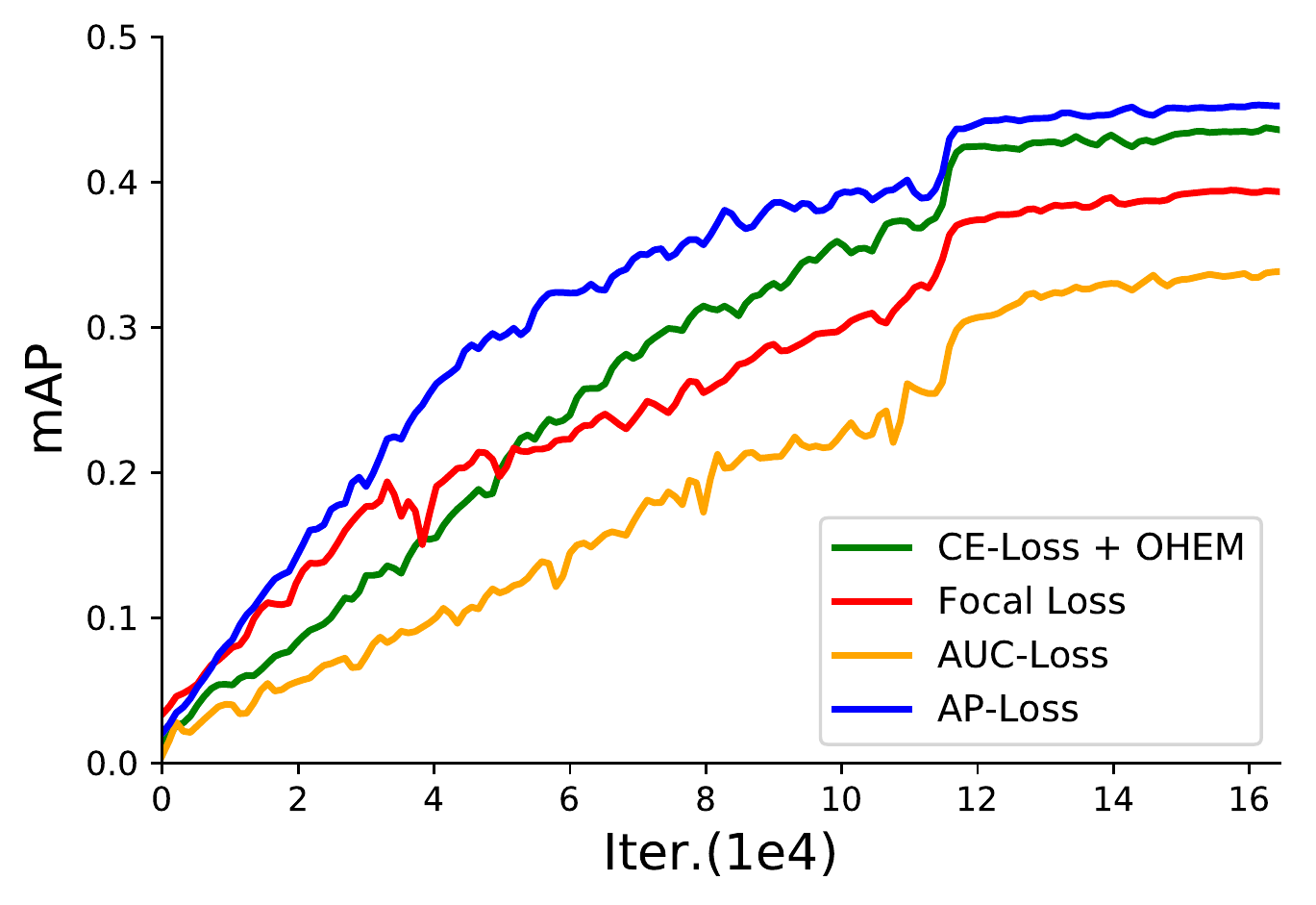}
\vspace{-4.5mm}
  \caption{Detection accuracy (mAP) on VOC2007 {\tt test} set.(Best viewed in color)}
\vspace{-2.5mm}
\label{fig-ssd_performance_curve}
\end{figure}

The results are shown in \autoref{base-loss-ssd} and \autoref{fig-ssd_performance_curve}. Note that the AP-loss outperforms all the other losses at both the final state and various snapshot time points. Together with the results on RetinaNet~\cite{lin2018focal} in Section 4.2.2 of the paper, we observe the robustness of the proposed AP-loss, which performs much better than the other competing losses on different datasets (\textit{i.e.} PASCAL VOC~\cite{everingham2015pascal}, MS COCO~\cite{lin2014microsoft}) and different detectors (\textit{i.e.} RetinaNet~\cite{lin2018focal}, SSD~\cite{liu2016ssd}). This demonstrates the effectiveness and strong generalization ability of our proposed approach.

\end{document}